\documentclass[a4paper,fleqn]{cas-dc}
\usepackage{float}
\usepackage{amsmath}
\usepackage{amssymb}
\usepackage{graphicx}
\usepackage{hyperref}
\usepackage[numbers]{natbib}

\begin{document}
\shortauthors{Y Zhu and ET Filipov}
\shorttitle{Simulation for Electro-Thermal Origami}
\title{Rapid Multi-Physics Simulation for Electro-Thermal Origami Systems}

\author[1]{Yi Zhu}[orcid=0000-0001-9667-834X]
\ead{yizhucee@umich.edu}
\credit{Department of civil and environmental engineering, University of Michigan}
\address{GG Brown Building, 2350 Hayward Street, Ann Arbor, 48109, MI, USA}

\author[2]{Evgueni T. Filipov}[orcid=0000-0003-0675-5255]
\ead{filipov@umich.edu}\credit{Department of civil and environmental engineering, University of Michigan}
\address{Department of civil and environmental engineering, University of Michigan}

\cortext[cor1]{Evgueni T. Filipov}

\begin{abstract}
Electro-thermally actuated origami provides a novel method for creating 3-D systems with advanced morphing and functional capabilities. However, it is currently difficult to simulate the multi-physical behavior of such systems because the electro-thermal actuation and large folding deformations are highly interdependent. In this work, we introduce a rapid multi-physics simulation framework for electro-thermally actuated origami systems that can simultaneously capture: thermo-mechancially coupled actuation, inter panel contact, heat transfer, large deformation folding, and other complex loading applied onto the origami. Comparisons with finite element models validate the proposed framework for simulating origami heat transfer with different system geometries, materials, and surrounding environments. Verification of the simulated folding behaviors against physical electro-thermal micro-origami further demonstrates the validity of the proposed model. Simulations of more complex origami patterns and a case study for origami optimization are provided as application examples to show the capability and efficiency of the model. The framework provides a novel simulation tool for analysis, design, control, and optimization of active origami systems, pushing the boundary for feasible shape morphing and functional capability. 
\end{abstract}

\begin{graphicalabstract}
\includegraphics[width=\linewidth]{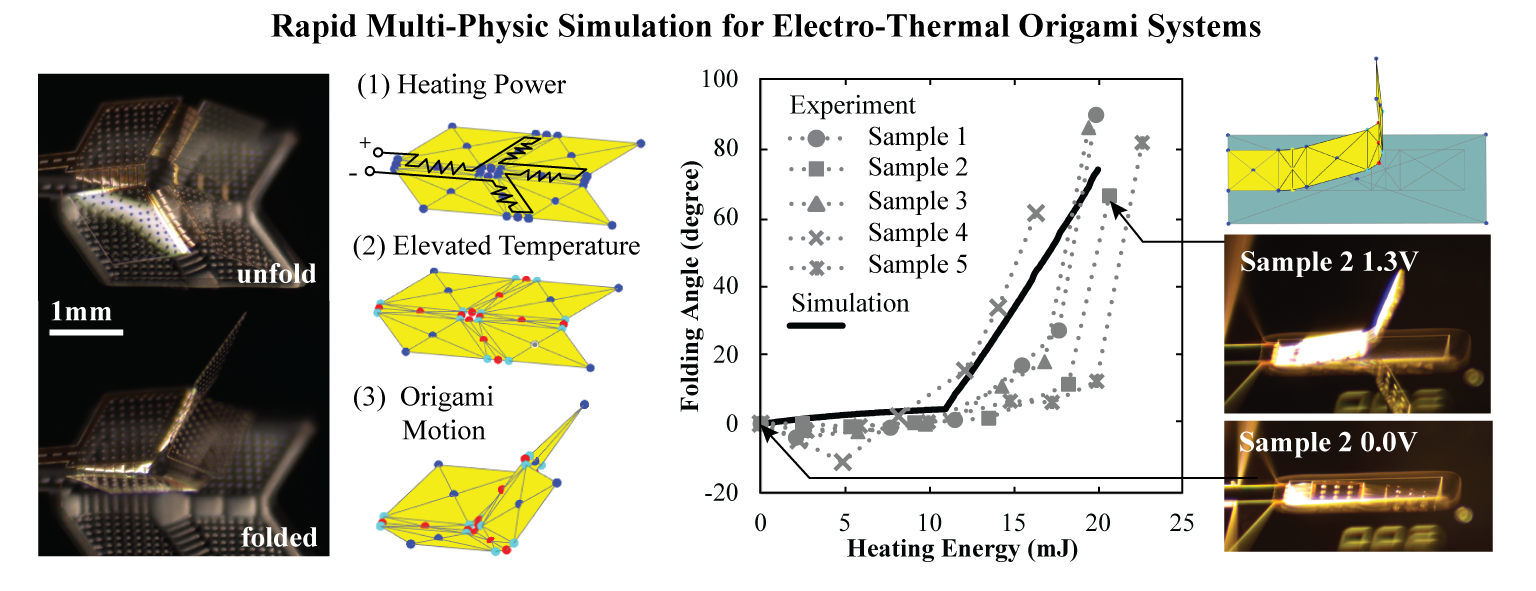}
\end{graphicalabstract}

\begin{highlights}
\item Rapid simulation framework for multi-physics actuation of electro-thermal origami.

\item New simplified heat transfer models calibrated and validated with finite elements.

\item Simulation for thermo-mechanical coupling verified with physical micro-origami.

\item Framework tested for simulating motion in complex systems and for optimization.

\item Open-access MATLAB package of the simulation framework is provided.

\end{highlights}

\begin{keywords}
micro-origami, electro-thermal actuation, origami heat transfer, bar-and-hinge model
\end{keywords}

\maketitle

\section{Introduction}

\textcolor{black}{Origami principles provide novel methods to create active systems that can self-assemble into complex 3-D configurations and can fold to achieve active functionalities \cite{Felton2014,Rus2018,Sareh2020}. The folding and unfolding motions of these active origami provide significant methods for designing engineered materials with tunable properties \cite{Overvelde2016,Boatti2017,Silverberg2015},  densely packing engineering systems\cite{Seymour2018,Filipov2015}, and  fabricating engineering systems with intricate 3D  geometries from a flat sheet \cite{Lang2018book,Na2015,NaurozePaulino2018,An2018}.  These active origami systems can be used for various applications such as meta-materials \cite{Schenk2013,Fang2018,Zhai2017,Li2019,Lv2014,Yang2017}, deployable space and building structures \cite{Lang2016,Filipov2019,Kaddour2020}, biomedical micro-grippers \cite{Zhu2020AFM,Breger2015}, energy absorption devices \cite{Ma2014,Xiang2020}, robotic systems \cite{Felton2014,Rus2018,Kamrava2018,Liu2021} and more.} 

More specifically, active origami systems are well suited for creating functional micro-scale devices because conventional micro-fabrication processes tend to produce planar systems and have difficulties to directly build structures with intricate 3-D geometries \cite{Rogers2016}. With the help of active origami systems, these micro-scale devices can be first fabricated within a 2-D plane and then folded into the desired 3-D geometry using micro-actuators \cite{Zhu2020AFM,Na2015}. 

\begin{figure*}[t]
\begin{center}
\includegraphics[width=\linewidth]{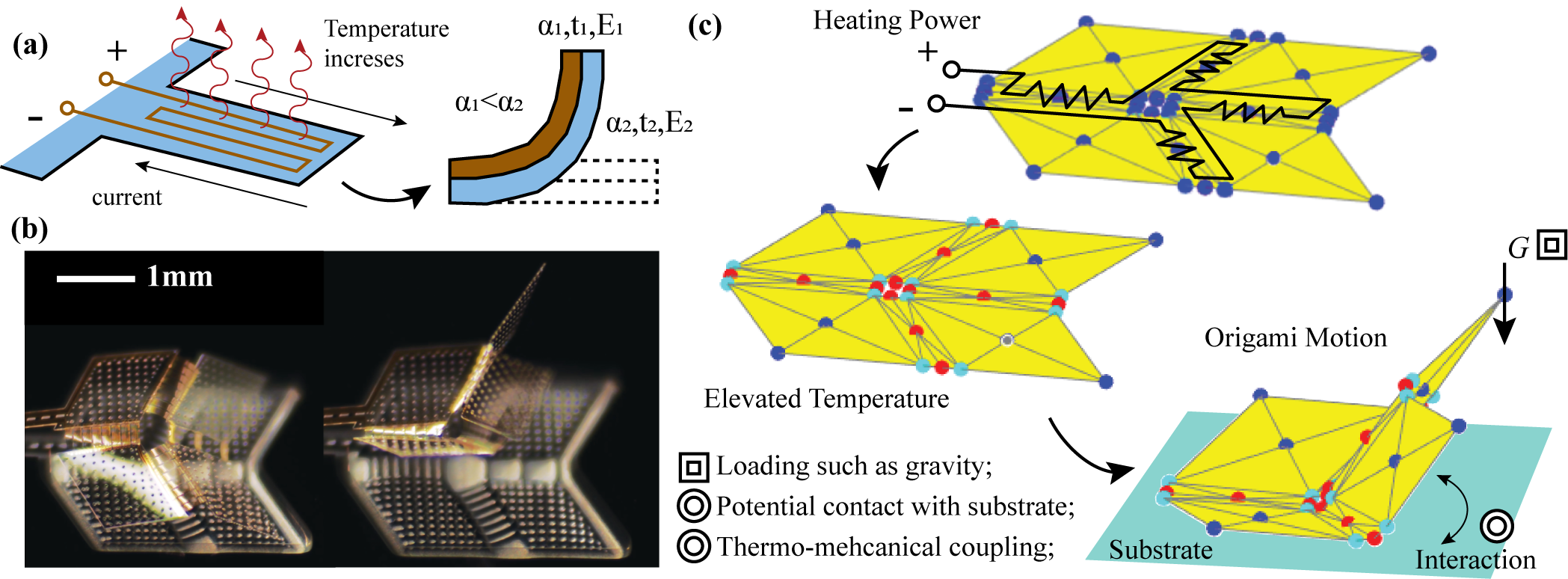}
\end{center}
\caption{(a) Actuation mechanism for the electro-thermal origami systems: Joule heating generated from the heating wire elevates the crease temperature and the system folds because the bottom layer expands more than the top layer; (b) A physical Miura-ori pattern micro-origami enabled by electro-thermal actuation. The structure is made using a fabrication process proposed in \cite{Zhu2020AFM}; (c) This work introduces a multi-physics based framework for simulating this genre of active origami systems. \textcolor{black}{The framework can simultaneously simulate compliant creases\cite{Zhu2020JMR}, inter-panel contact\cite{Zhu2019PRSA}, large deformation folding\cite{Liu2017PRSA}, and thermo-mechanical coupling (this work) of active origami.}}
\label{figure_Intro} 
\end{figure*}

Over the past decades, a number of active  origami systems have been built and tested. These origami are actuated with active materials or responsive systems, such as hydrogels \cite{Na2015,Yoon2014,Kang2019}, metallic morphs with residual stress \cite{Leong2009,Leong2008,Bassik2009}, shape memory polymers \cite{Felton2014,An2018}, magnetic active systems \cite{Shaar2015,Iwase2005}, etc. Among these different systems, thermally active materials demonstrate great potential in building practical origami devices \cite{Na2015,Leong2009,Leong2008} \textcolor{black}{, because thermally active material can achieve high energy efficiency and provide high work density \cite{Yang2020}.} Moreover, by designing proper electric circuits to control the heaters, one can control the folding motion of origami creases separately to achieve complex motions and functions \cite{Zhu2020AFM,Jager2000}. Figure~\ref{figure_Intro} (a) to (c) demonstrate how electro-thermal actuators can enable origami systems to achieve controllable folding at the micro-scale. When passing current through the patterned metallic heaters, Joule heating will elevate the temperature of the crease regions. If the creases are made with a bi-material morph that has two different thermal expansion coefficients ($\alpha$), the changing temperature will generate bending motion because one layer will expand more than the other. Figure~\ref{figure_Intro} (b) demonstrates a Miura-shape micro-origami enabled by this type of electro-thermal actuation. 

However, simulation of electro-thermally actuated active origami has remained substantially limited. Current simulation methods for origami are mostly kinematics-based \cite{Tachi2009Rigid,Tachi2010} or mechanics-based \cite{Liu2017PRSA,Liu2016Merlin} and have incomplete abilities to capture the multi-physics based actuation. As demonstrated in Fig.~\ref{figure_Intro} (c), properly capturing the folding motion is indeed difficult, as it requires simultaneously simulating the coupled electro-thermal heating, the interaction between heat dissipation and large deformation folding, the potential contact between substrate and origami panels, and other loading effects such as gravity. This difficulty in simulation has limited design methods for active origami to mostly trial and error approaches \cite{Na2015,Zhu2020AFM}. Furthermore, without efficient simulation it is difficult to optimize designs or to construct control protocols for these active origami systems. Therefore, the primary goal of this work is to answer the question: \textit{How can we accurately and efficiently simulate the folding motion of electro-thermally active origami?}

To resolve this problem, we created a novel simulation framework based on the bar and hinge representation of origami \cite{Schenk2010}. The proposed framework can capture electrically generated local heating, thermally induced crease curvature, thermo-mechanically coupled large folding, contact induced panel interaction, and other loading such as gravity (Fig.~\ref{figure_Intro} (c)). More significantly, this work also tested the proposed framework against physical active origami devices built with  micro-fabrication. Good agreement between the simulated folding behaviors and the measured ones demonstrates the validity of the framework. This work provides the much needed multi-physics based simulation method for building practical origami inspired devices in a multi-physical environment. Moreover, the framework proposed in this work is applicable to active origami at multiple length scales and can be used generally to simulate other geometrically planar active systems. 

This paper is organized as follows. First, to motivate this work we give a brief introduction of current origami simulation methods, pointing out why these methods have difficulties capturing the thermo-mechanically coupled folding motion of origami systems. Next, we introduce the proposed simulation framework in detail. We describe the three major steps of the framework and introduce the modeling techniques involved. The fourth section of this paper presents a series of calibration tests used to fine tune the model parameters and to check the accuracy of the new heat transfer model against high-fidelity finite element simulations. Next, we verify the performance of the proposed simulation framework by comparing it against physical experiments. Finally, the paper presents three practical examples to demonstrate the broad capability, efficiency, and usefulness of this rapid simulation framework. 

A code package implementing the rapid simulation method is made available at GitHub: \url{https://github.com/zzhuyii/OrigamiSimulator}. The online version is continuously updated with additional origami simulation capabilities.


\section{Related Work}

In this section, we introduce the state-of-the-art simulation methods for origami systems. More specifically, we will focus on why currently available methods have difficulties in simulating the electro-thermo-mechanically coupled actuation of origami systems. 


One pioneering method to study the folding motion of origami is the rigid folding algorithm proposed by Tachi \cite{Tachi2009Rigid,Tachi2009Quad,Tachi2010}. The method solves the folding motion of an origami considering the kinematic constraints, such as Kawasaki's theorem \cite{Kawasaki1988}, and tracks the motion using Euler's methods. This method was used to plan the folding motion of origami robots such as those in \cite{An2018,Hawkes2010}. However, the model is kinematics-based and thus does not have the capability to consider the influence of gravity, thermal loading, and folding mechanics which are essential for capturing the realistic behaviors of electro-thermal origami systems.

The bar and hinge model is a mechanics-based model that represents an origami with extensional bar elements and rotational springs elements \cite{Schenk2010,Liu2017PRSA}. The bars are used to capture in-plane behavior such as panel stretching and shearing, while the rotational springs are used to capture the out-of-plane behaviors. The detailed formulation of these elements can be found in \cite{Liu2018Merlin2,Liu2017PRSA,Filipov2017}. The model can capture the mechanics and the stiffness of crease folding, panel bending, panel stretching and shearing within an origami. Recent advancements of the bar and hinge model have enabled it to simulate compliant creases \cite{Zhu2019IDETC,Zhu2020JMR} and panel contact \cite{Zhu2019PRSA} within active origami systems. \textcolor{black}{The models demonstrated in \cite{Zhu2019IDETC,Zhu2020JMR,Zhu2019PRSA} provide the basis of the simulation framework introduced in this work. In section~\ref{Solving_equilibrium}, we will briefly introduce how to model the compliant creases and inter panel contacts.} However, these models do not consider the heat-transfer aspects of the origami and thus cannot be used to estimate the crease temperature and the resulting folding angle of creases. 

Finite elements (FE) based models have also been used to study the behaviors of origami structures \cite{Fang2018,Ma2014,Ma2018}, but have been limited mostly to the simulation of mechanical behaviors. Besides, the long computation time and extensive work required for building FE models have pushed origami researchers to search for faster and easier to use alternatives. Separately from origami, FE models have been used to study electro-thermal actuators for micro-electromechanical systems (MEMS) \cite{Hussein2016,Ozsun2009,Iamoni2014}. However, these FE models developed for MEMS cannot be directly applied to the simulation of electro-thermal active origami systems because typical MEMS devices produce only small deformations where the thermo-mechanical coupling can be neglected. In contrast, origami-inspired active systems experience large deformations and strong thermo-mechanical coupling during the folding process. Properly capturing this coupled behavior using the FE model would require re-meshing of the structure and its surrounding environment every iteration. This re-meshing adds extensive complexity and time to the simulation, and makes FE models unfeasible for the problem at hand. 

To overcome the limitations of the existing methods for simulating origami, we propose a novel rapid simulation framework that is built upon the bar and hinge model but has the capability to capture the multi-physics associated with the electro-thermal actuation. The framework can simulate the heat transfer process within origami systems as well as the coupling between the thermal loading and mechanical folding. This framework provides the much needed tool for the design and optimization of active origami systems.

\section{Simulation Framework for Active Origami}

\begin{figure}[t]
\begin{center}
\includegraphics[width=\linewidth]{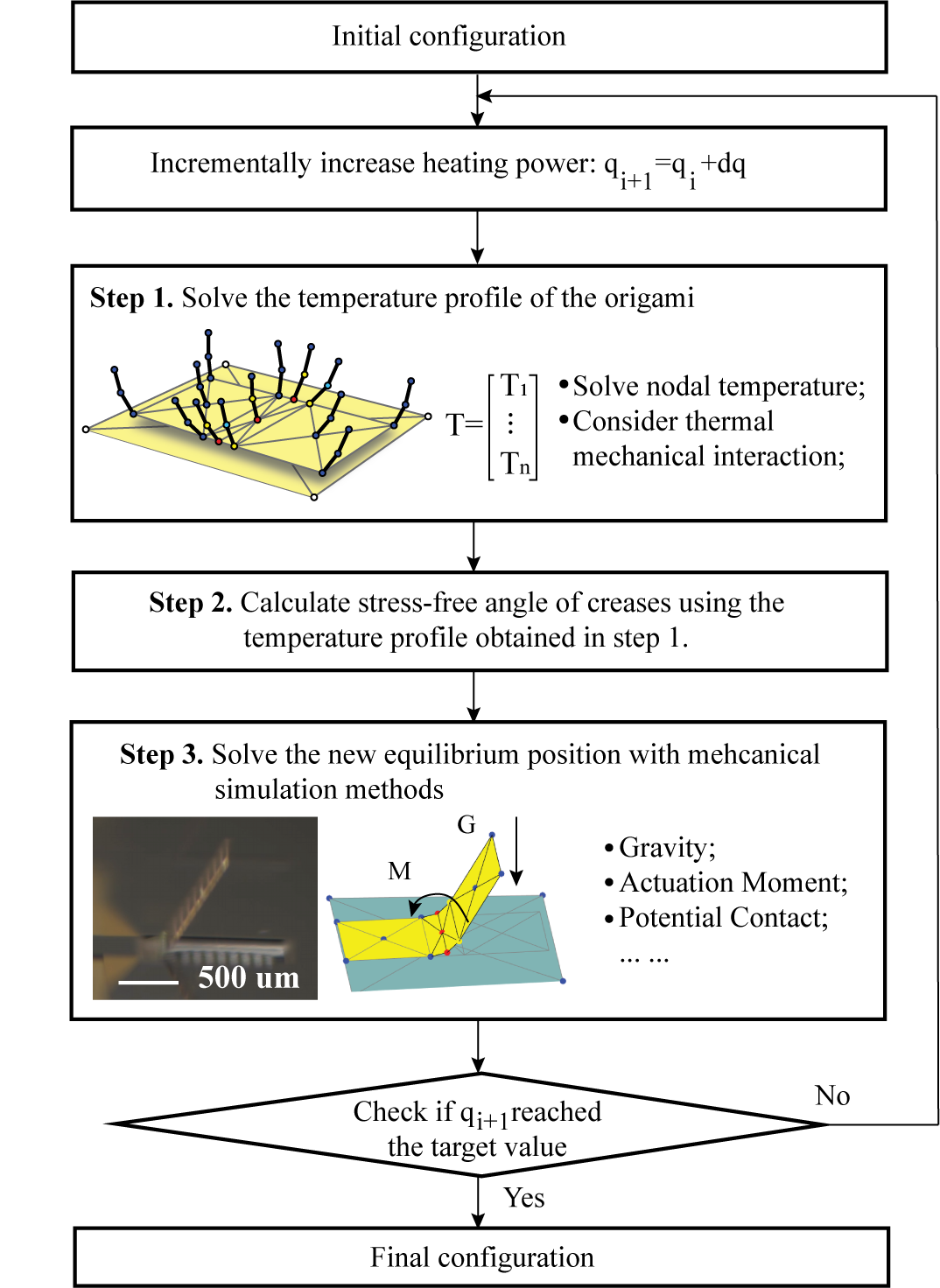}
\end{center}
\caption{Flowchart for the rapid simulation framework for electro-thermally actuated origami systems. }
\label{figure_SimFrame} 
\end{figure}

In this section, we introduce the proposed rapid simulation framework for electro-thermal micro-origami systems, which is based on a simple but effective bar and hinge model. The simulation framework has three major steps: (1) simulate the temperature profile under applied heating power, (2) calculate the stress-free angle of creases based on the elevated temperature, and (3) solve the new equilibrium position of origami system (Fig.~\ref{figure_SimFrame}). In the following subsections, we will describe these steps in detail.

\subsection{Step 1: Solving the heat transfer problem}

\begin{figure*}[t]
\begin{center}
\includegraphics[width=\linewidth]{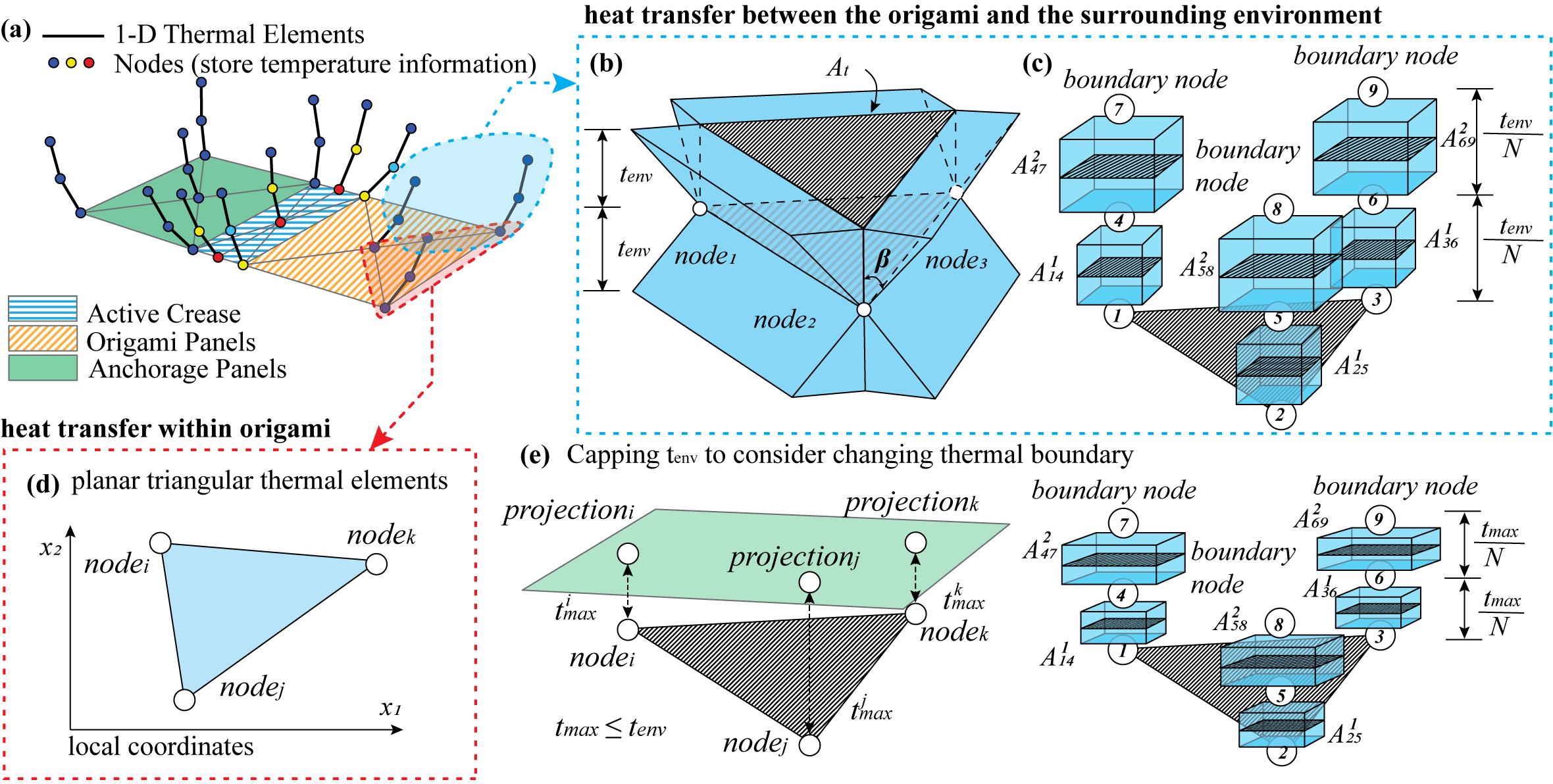}
\end{center}
\caption{(a) A schematic representation of the bar and hinge model for simulating the heat transfer problem; (b) The surrounding substance  considered for the heat loss to the environment; (c) One dimensional bars and additional nodes to represent the thermal conductivity between the structure and the surrounding environments. The bar cross section area is derived such that the total volume of the surrounding substance is preserved; (d) Planar triangular thermal elements are used to capture heat transfer within the origami structure; (e) When there is a nearby thermal boundary and $t_{max}\leq t_{env}$, $t_{max}$ is used to calculate the conductivity for structure-environment heat loss.}
\label{figure_BandHthermal} 
\end{figure*}

First, we focus on how to capture the elevated temperature of the origami systems under applied heating power (see Fig.~\ref{figure_BandHthermal}). This heat transfer problem has two major components: the heat transfer within the origami structure, and the heat transfer between the structure and the surrounding environment (Fig.~\ref{figure_BandHthermal} (a) to (d)). For the heat transfer problem within the origami structure, we use the existing nodes in the bar and hinge model to store the temperature information and use the planar triangular (T3) thermal elements to generate the conductivity matrix (see Fig.~\ref{figure_BandHthermal} (d)). This elements are a convenient choice because the bar and hinge model naturally provides triangulated meshing of the origami surface. Next, to consider the heat transfer between the origami and the surrounding environment, we introduce additional nodes to represent the temperature of the surrounding environment (e.g. air, water) near the surface of these active systems. We simplify the heat transfer between the structure and the surrounding environment as a 1-D thermal conduction problem and use multiple bars to represent the thermal conductivity between the structure and the surrounding environment. Generally speaking, simulating the heat transfer as a conduction problem is a valid assumption for small-scale systems, because convection is not significant. For most devices fabricated with MEMS processes, this is a valid assumption \cite{Hussein2016,Ozsun2009}. In the remainder of this subsection, we will introduce how to calculate the model parameters based on the system geometry and the material properties to effectively capture the heat transfer between the origami and the surrounding environment. 

A chain of nodes and bars is added to each node of the origami to represents the surrounding environment (Fig.~\ref{figure_BandHthermal} (a) to (c)). This chain of nodes and bars captures the heat transfer between the structure and the surrounding environment as an 1-D thermal conduction problem. We assume that the end nodes (farthest away from structure) are at room temperature (RT). The conductivity of bars in this chain representing the environmental substance column can be calculated as:

\begin{equation}\label{eq_barThermal}
k_{ij}=\frac{k_{mat}A_{ij}}{L_{ij}},
\end{equation}

\noindent where $k_{mat}$ is the thermal conductivity of the surrounding environment, $L_{ij}$ is the bar length between node $i$ and node $j$, and $A_{ij}$ is the cross section area of the bar (see Fig.~\ref{figure_BandHthermal} (c)).  

Figure~\ref{figure_BandHthermal} (b) and (c) introduce how to calculate the bar area $A_{ij}$ connecting node $i$ and node $j$. The equation for assigning the bar area $A_{ij}$ is derived by preserving the total thermal conductivity of the bulk material of the surrounding environment. Consider a triangular heating panel as shown in Fig.~\ref{figure_BandHthermal} (b) and assume that at thickness $t_{env}$ away from the panel, the surrounding environment is at room temperature. To model the effect that a larger volume of surrounding substance gets heated as the distance from the heating panel increases, we introduce the dissipation angle $\beta$. Then the total volume of surrounding substance at one side of the panel is the sum of four triangular prisms as pictured in Fig.~\ref{figure_BandHthermal} (c). We further introduce $N$, the total number of bars in the chain, to convert this gradually changing cross section into $N$ bars with different constant cross sections $A^{k}_{ij}$ (where the superscript $k$ indicates that this area is for layer $k$). To match the total thermal conductivity for layer $k$, we need to preserve the total volume and thus we have:

\begin{equation}\label{eq_barAirStructure}
\frac{3t_{env}A^{k}_{ij}}{N}=2\frac{t_{env}}{N}(A_t+L_{sum}t_{env}\tan{\beta}(\frac{k-0.5}{N})),
\end{equation}

\noindent where $A_t$ is the area of the triangle heating panel and $L_{sum}$ is the length of the perimeter of the heating panel. By rearranging the equation, we obtain the expression of the cross section of bars for each layer in the chain as:

\begin{equation}\label{eq_barAAir}
A^{k}_{ij}=\frac{2}{3}(A_t+L_{sum}t_{env}\tan{\beta}(\frac{k-0.5}{N})).
\end{equation}

Tuning the value of $N$, $\beta$, and $t_{env}$ allows us to capture the heat transfer between the origami and the surrounding environment properly. The effects of different combinations of these parameters are studied closely in the following section. 




After solving the thermal conductivity of different bars, we can assemble the global thermal conductivity matrix and solve the nodal temperature as $\boldsymbol{T}=\boldsymbol{K}^{-1}\boldsymbol{q}$, where $\boldsymbol{T}$ is the nodal temperature vector, $\boldsymbol{K}$ is the global conductivity matrix, and $\boldsymbol{q}$ is the input nodal heating power. 

One major advantage of the proposed bar and hinge framework is that it can simulate the thermo-mechanical coupling by varying the $t_{env}$ parameter based on the geometric configuration of origami. Figure~\ref{figure_BandHthermal} (d) shows how  the thickness of surrounding substance is capped to consider the changing thermal boundary. When a triangular heating panel gets close to a thermal boundary, the distances between the three nodes of the heating panel and the planar thermal boundary are calculated, and the average $t_{max}=(t_{max}^i+t_{max}^j+t_{max}^k)/3$ is used to cap $t_{env}$. When $t_{max}\leq t_{env}$, $t_{max}$ is used to calculate the conductivity for structure-environment heat loss. When $t_{env}\leq t_{max}$, the boundary has limited effects on the heat transfer problem, and $t_{env}$ is used to calculate the conductivity for structure-environment heat loss. In the verification section, we will show that this simple method allows the refined bar and hinge model to capture the thermo-mechanically coupled folding behavior of the active origami effectively. While this thermal model is built specifically for small-scale origami structures, we envision that it can be extended to model thermal behavior of other thin origami and kirigami inspired structures. 

\textcolor{black}{When following the flowchart on Fig.~\ref{figure_SimFrame} for solving the heat transfer problem of active origami system, it is important to select an appropriate  step size for heating $dq$. If the heating step $dq$ is too large,  step 3 of the flowchart can have convergence problems. In general, we suggest controlling the heating step $dq$ to be relatively small so that step 3 can converge within 5 to 10 Newton iterations. A smaller $dq$ value will require more increments to finish the simulation and thus makes the wall-clock time longer. However, this longer run time may be preferred because it stabilizes the simulation and ensures easier convergence of the problem.}

\subsection{Step 2: Solving the stress-free angle}

Next, we compute the stress-free angle of the folding creases based on the nodal temperature profile obtained in step 1. In this work, we consider the crease region to be made of bi-material morph components that consist of two layered materials with different coefficients of thermal expansion. Because one of the material layers expands more when heated, the bi-material morph will develop curvature and folding in the crease region (Fig.~\ref{figure_Intro} (a)). \textcolor{black}{This is a common actuator design which has been used in many previous research efforts \cite{Na2015,Zhu2020AFM,Kang2019}. Furthermore, we assume that the panels of the origami will remain near to rigid, and that the elevated temperature will not introduce significant deformation in the panels that interferes with the crease folding. This is a reasonable assumption because most active origami system have  panels that are rigid when compared to the creases, and also the panels remain passive to the applied stimulus \cite{Zhu2020AFM,Na2015,Kang2019,Felton2014}}. Timoshenko's bi-material morph model \cite{Timoshenko1925} is used to calculate the stress-free curvature of the creases based on the elevated crease temperature. This analytical model assumes that both the section level structural response and the material response are linear. Though simple, the model gives a good prediction of the curvature and rotation of the bi-material morphs, and its accuracy has been demonstrated in previous research \cite{Zhu2020AFM,Na2015,Pezzulla2015}. The curvature of the bending actuator under a given elevated temperature $\delta T$ is calculated as:

\begin{equation}\label{eq_timo}
\kappa=\frac{6(\alpha_1-\alpha_2)(1+m)^2 \delta T}{(t_1+t_2)[3(1+m)^2+(1+mn)(m^2+\frac{1}{mn})]},
\end{equation}

\noindent where $\alpha_1$ and $\alpha_2$ are the thermal expansion coefficient of the two materials, $E_1$ and $E_2$ are the Young's moduli of the two materials, $t_1$ and $t_2$ are the thicknesses of the two materials, $m=t_2/t_1$, and $n=E_2/E_1$. The stress-free angle of the folding crease with length $l$ can be next calculated as: $\theta=\kappa l$. The updated stress-free angle of folding creases will be used in the mechanics formulation in step 3 to determine the new equilibrium shape of the origami systems.

\subsection{Step 3: Solving the equilibrium position}\label{Solving_equilibrium}

\begin{figure}[t]
\begin{center}
\includegraphics[width=\linewidth]{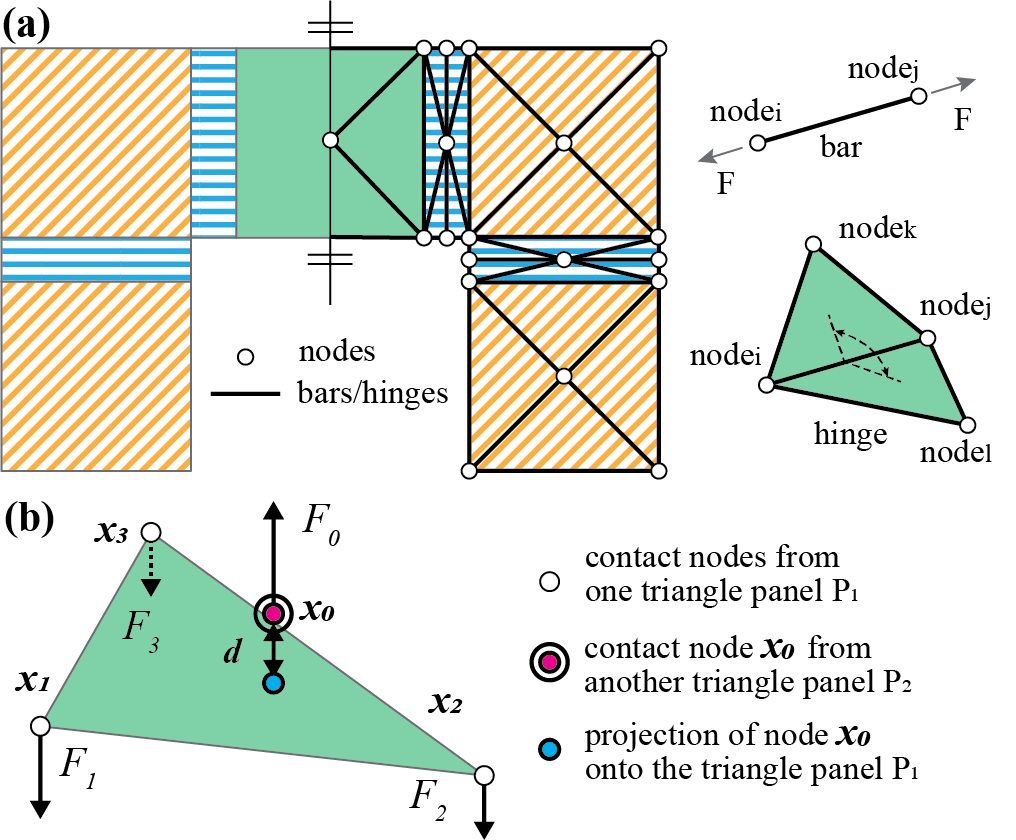}
\end{center}
\caption{(a) The bar and hinge model for simulating large-deformation folding motions for active origami structures; (b) Global contact between a node $x_0$ and a triangular panel is captured using the contact model proposed in \cite{Zhu2019PRSA}.}
\label{figure_BH} 
\end{figure}

Here, the bar and hinge model is used to simulate the new equilibrium position of the origami system. In the bar and hinge model, the origami is represented using extensional bar elements that capture the in-plane panel shearing and stretching, and rotational springs (hinges) that capture the out-of-plane crease folding and panel bending (Fig.~\ref{figure_BH}). The total potential strain energy stored within the origami system can be expressed as the following:

\begin{equation}\label{eq_potential}
U (x)=U_{bar} (x)+U_{spr} (x)+U_{contact}(x),
\end{equation}

\noindent where $x$ is the nodal coordinates, and $U_{bar}$, $U_{spr}$, and $U_{contact}$ are the potential energies from stretching of bar elements, folding of rotational springs, and contact of system components. By the formulation of principle of stationary potential energy, searching for the configuration with the local minimum potential will yield a configuration that is in equilibrium. In this work, we adopt the model proposed in \cite{Zhu2020JMR} as a basis for simulating the large folding mechanisms of active origami. Readers can refer to \cite{Zhu2020JMR,Liu2017PRSA,Filipov2017} for detailed formulation of the bar and hinge model.

\textcolor{black}{Contact within the origami model can be detected based on the distance $d$ between nodes and surfaces in the modele (origami panels or substrate). To simulate the contact interactions, a contact penalty function is included into the total potential of the system as indicated in Eqn.~(\ref{eq_potential}). The contact potential function is formulated using the distance $d$ between a triangulated panel that represents a surface and a separate node from another panel as:}

\begin{equation}\label{eq_contact}
U(d)=
\begin{cases}
\begin{matrix}
\begin{matrix}
k_e\{ln[sec(\dfrac{\pi}{2}-\dfrac{\pi d}{2d_0})]-
\\
\dfrac{1}{2}(\dfrac{\pi}{2}-\dfrac{\pi d}{2d_0})^2 \} 
\end{matrix}
& (d \leqslant d_0) 
\\
0 & (d>d_0),
\end{matrix}
\end{cases}
\end{equation}

\noindent where $d_0$ is the threshold at which the contact initiation and disengagement occur (Fig.~\ref{figure_BH}). When $d>d_0$ the contact potential gives zero output and the contact induced potential is only considered when we have $d\leq d_0$. This contact potential acts as a penalty function that generates a large potential value (approaching infinity) when the distance between the contacting point and triangle is small (when $d$ is approaching zero). Due to the presence of this penalty function, the algorithm is discouraged from having closely spaced panels when searching for the local minimum in total potential which thus prevents the panels from penetrating each other. The detailed formulation for capturing the contact behavior within the bar and hinge model can be found in \cite{Zhu2019PRSA}. 

Together, the simulation framework allows us to capture: the global temperature distribution induced by electro-thermal heating, crease curvature resulting from the elevated temperature, interaction between heating dissipation and the large deformation folding, mechanical deformation from gravity and other loading, and contact induced interactions within the origami systems.

\begin{figure*}[t]
\begin{center}
\includegraphics[width=0.95\linewidth]{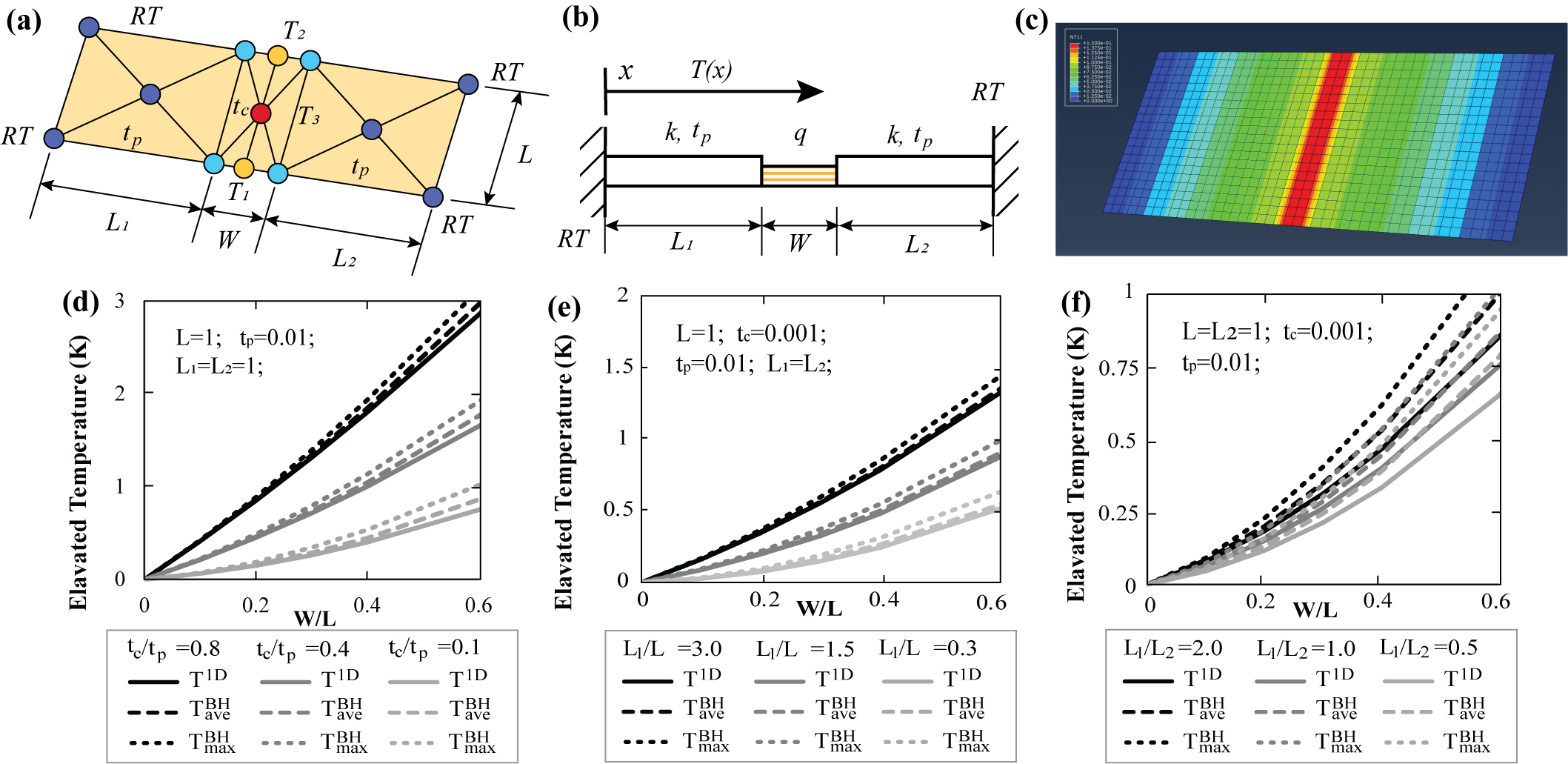}
\end{center}
\caption{Verification of the proposed simulation for heat transfer within the origami structure. (a) Model set up and the geometry of a two-panel origami system; (b) The system can be modeled as an 1-D heat transfer problem with existing analytical solution; (c) A FE simulation with $W/L=0.2$, $t_{c}/t_{p}=0.1$, and $L_1/L=1$ demonstrates the 1-D nature of the problem; (d-f) Relations between the temperatures at the crease region simulated by the bar and hinge model ($T_{max}^{BH}$ and $T_{ave}^{BG}$) and the temperature for an equivalent 1-D heat transfer problem ($T^{1D}$). The relations are shown with respect to W/L for different cases of the (d) $t_c/t_p$ ratio, (e) $L_1/L$ ratio, and (e) $L_1/L_2$ ratio.}
\label{figure_VerificationWithin} 
\end{figure*}

\begin{figure*}[t]
\begin{center}
\includegraphics[width=0.95\linewidth]{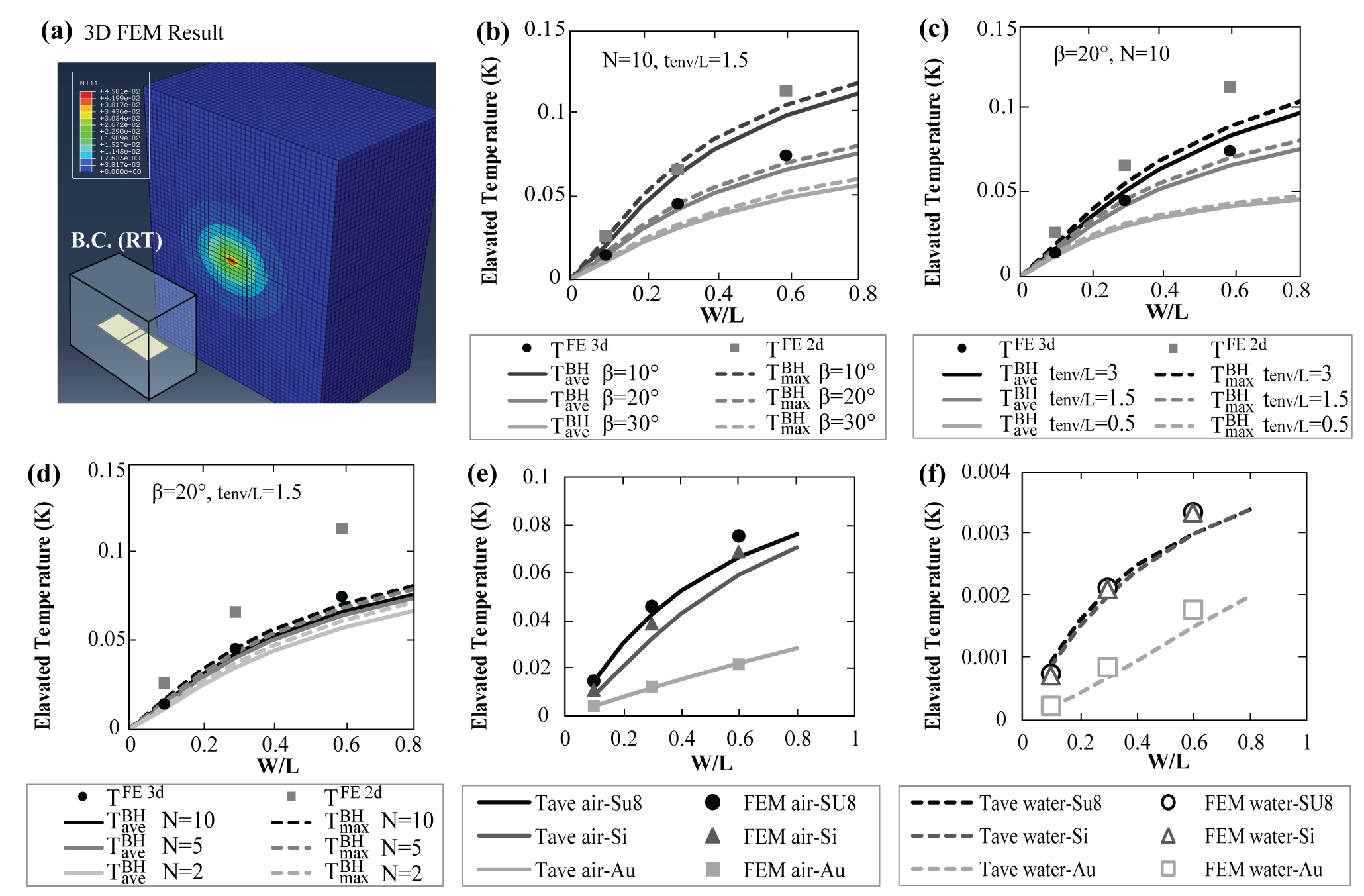}
\end{center}
\caption{Calibration of the heat transfer between the origami and the surrounding environments. (a) A 3-D FE simulation of the system ($W=0.3,L=0.1$); (b-d) Relations between the temperatures at the crease region simulated by the bar and hinge model ($T_{max}^{BH}$ and $T_{ave}^{BH}$ ) and by the 2-D and 3-D finite element models ($T^{FE2d}$ and $T^{FE3d}$). The relations are shown with respect to W/L for different  (b) dissipation angles ($\beta$),  (c) environmental substance layer thickness ($t_{env}$), and (d) numbers of discretizing layers; (e-f) The temperature change with respect to $W/L$ for different structure materials when the surrounding environment is (e) air and (f) water. }
\label{figure_VerificationLoss} 
\end{figure*}

\section{Calibration and Parameter Selection}

This section studies the behavior of the proposed simulation framework for the heat transfer problems  (step 1 of the framework) and verifies its performance. The validity of the models for solving crease curvature (step 2) \cite{Zhu2020AFM,Na2015,Pezzulla2015} and for solving the large deformation origami motion (step 3) \cite{Zhu2020JMR,Zhu2019PRSA} have been studied previously. Moreover, this section also demonstrates how we select the model parameters (thickness of surrounding environment at room temperature  $t_{env}$, number of layers $N$ , and dissipation angle $\beta$) used to tune the behavior of the heat transfer between the structure and the surrounding environment. Additional studies further verify that the chosen set of parameters are applicable to a wide range of combinations of geometries and material properties. 

\subsection{Heat transfer within origami structures}

We first study the validity of the proposed bar and hinge model for simulating the heat transfer problem within the origami structure. Because our model uses the well established planar triangular (T3) thermal elements, the performance is guaranteed when fine meshing is used. However, the proposed bar and hinge model only uses a coarse mesh so it is necessary to determine how much error is introduced in the temperature prediction. Figure~\ref{figure_VerificationWithin} (a) shows the setup for a two-panel origami with one heating crease system, which we use to explore the performance of the model. We set the far ends of the two panels to be at room temperature ($RT=0$), and apply a uniform body heat of magnitude $q=10$ in the center crease. The planar geometry of the panels is defined using $L$, $L_1$, and $L_2$, while the crease is defined using $L$ and $W$. The thickness of the panels is $t_p$, and the thickness of the crease is $t_c$. This system can be effectively modeled as an 1-D heat transfer problem with available analytical solutions.

%
%
%
%

We assume a non-dimensionalized material thermal conductivity $k=1$ for all the verification studies in this subsection. A FE simulation of the two panel origami shows the 1-D characteristic of this heat transfer problem (Fig.~\ref{figure_VerificationWithin} (c)).  Four non-dimensional parameters are studied for this comparison and they are: (1) $W/L$ indicating the ratio between the width of the creases to the length of the creases; (2) $t_c/t_{p}$ indicating the ratio between the thickness of the creases and the thickness of the panels; (3) $L_1/L$ where $L_1=L_2$ indicating the aspect ratio of the panels; and (4) $L_1/L_2$ indicating different lengths of the two panels. 

The proposed bar and hinge model cannot capture the 1-D heat transfer feature perfectly because the three nodes on the center line of the crease give different temperature predictions. However, as we will show, these predictions are close to the theoretical 1-D heat transfer solution. For the verification, we compare the difference between the maximum simulated temperature $T_{max}^{BH}=T_3$, the average simulated temperature $T_{ave}^{BH}=(T_1+T_2+T_3)/3$, and the analytical solution of the 1-D heat transfer problem $T^{1D}$. 

First, we assume that the geometry of the two panels is identical by setting $L_1=L_2$ and study the performance of the model for different crease versus panel thicknesses ($t_c/t_p$ - Fig.~\ref{figure_VerificationWithin} (d)) and for different length versus width of the structure ($L1/L$ Fig.~\ref{figure_VerificationWithin} (e)). The prediction from the proposed bar and hinge framework closely matches the analytical solution and captures the identical trends for different crease dimensions ($W/L$). Both curves of $T_{max}^{BH}$ and $T_{ave}^{BH}$ match the analytical solution well and there is limited difference between the value of $T_{max}^{BH}$ and $T_{ave}^{BH}$. Next, we break the assumed the symmetry of the system and study the performance of the model when the two panels have different geometries ($L_1/L_2$ Fig.~\ref{figure_VerificationWithin} (f)). Similarly, the bar and hinge model can capture the trends well. Although the maximum temperature of the crease is slightly overestimated, the average temperature of the crease is about the same as the analytical solution. These results indicate that the non-uniformity of the coarse bar and hinge mesh do not introduce a significant error regardless of the system geometries.

\subsection{Heat transfer between origami structures and surrounding environments}


To test the performance of the model for simulating heat transfer between the structure and the surrounding environment, we need to study how the three model parameters affect the model performance, namely: thickness of surrounding substance between the structure and the point at room temperature  $t_{env}$, the number of discretizing layers $N$ , and the dissipation angle $\beta$. We vary the combinations of these three parameters and compare the results from the bar and hinge model to a 2-D FE simulation and a 3-D FE simulation of the same two panel origami system (see Fig.~\ref{figure_VerificationLoss} (a) for the set up). The 3-D FE simulation represents the realistic behavior of the system because it is able to consider heat loss in all directions. On the other hand, the 2-D FE simulation result is provided as an upper bound. We assume the surrounding environment to be air (with thermal conductivity of $k_{air}=0.026 W\cdot m^{-1} \cdot K^{-1}$ \cite{Huang1999}) and assume the structure is made from an SU-8 polymer (with thermal conductivity of $k_{SU8}=0.3 W\cdot m^{-1} \cdot K^{-1}$ \cite{MicroChemSU8}), which is a common polymer material for MEMS devices.

The comparisons of the results for different combinations of the three model parameters are summarized in Fig.~\ref{figure_VerificationLoss} (b), (c), and (d). Figure~\ref{figure_VerificationLoss} (b) suggests that our model is sensitive to the selection of the dissipation angle $\beta$. This $\beta$ parameter allows us to consider the heat loss in different directions and is determined by the geometric characteristics of the system. The results presented suggest that $\beta=20^{\circ}$ matches the 3-D FE simulation well for a typical single crease origami system, and thus this value is selected for further simulations performed in the paper. Figure~\ref{figure_VerificationLoss} (e) shows that a value of $t_{env} = 1.5 L$ is needed for the thickness of the surrounding substance such that the proposed model can properly capture the surrounding environment. Figure~\ref{figure_VerificationLoss} (f) shows that the simulation is not sensitive to the number of discretizing layers $N$, once $N$ is sufficiently large ($5\leq N$). The results presented in Fig.~\ref{figure_VerificationLoss} show that when the proper combination of model parameters are selected ($\beta=20^{\circ}$; $t_{env} = 1.5 L$; $N=10$), the proposed bar and hinge model can capture heat transfer between the origami and the surrounding environment with reasonable accuracy. The model captures the trends with respect to the geometry of the system (W/L ratio) and provides a reasonable approximation of the actual temperature changes in the 3-D system.  

Finally, we check the performance of the proposed model for different material combinations (Fig.~\ref{figure_VerificationLoss} (e) and (f)). We select air and water as the surrounding environments and use Su-8, Si, and Au as the materials of origami structures (these environments/materials are common for active origami systems). We use the same model parameters as those chosen through the parametric study in  Fig.~\ref{figure_VerificationLoss} ($\beta=20^{\circ}$; $t_{env} = 1.5 L$; $N=10$). The model shows a good match with the 3-D FE simulation regardless of the type of environmental substance and the type of material that is used for the origami structure. Thus the above model parameters are a reasonable choice for simulating different structure and environment combinations.

\section{Experimental Verification}

In this section, we will verify the performance of the proposed simulation framework by comparing the prediction of the folding motion from the model to physical micro-origami systems. The two cases to be studied are: a single-crease origami with a folding threshold and a double-crease origami with thermo-mechanical coupling. 

\subsection{Single-crease origami with a folding threshold}

\begin{figure}[t]
\begin{center}
\includegraphics[width=\linewidth]{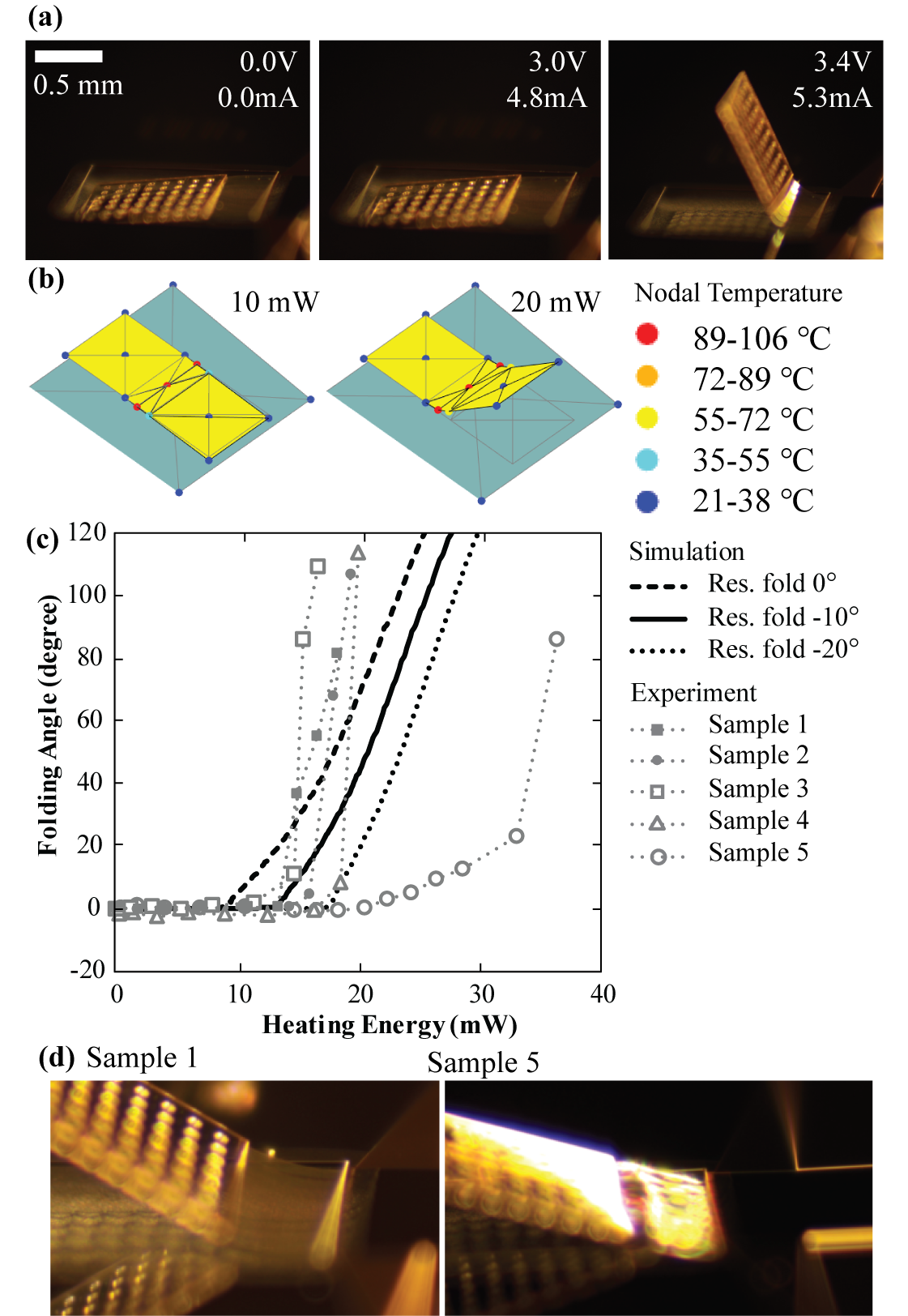}
\end{center}
\caption{Comparison of the simulated results between the bar and hinge model and the measured folding behavior of a single-crease micro-origami with 400 ${\mu}$m long actuators. (a) Folded geometry of the single-crease micro-origami at different voltage inputs; (b) Simulated folding process of the single-crease micro-origami with the proposed framework; (c) Comparison of the folding angle to input power curves between the simulation and the measurements; (d) Local defects such as the wrinkling in Sample 5 sometimes result in experimental outliers. See supplementary video Verification 1 for a comparison of the simulated animation and the recorded motion of the physical devices.}
\label{figure_SingleCrease} 
\end{figure}

\begin{figure*}[t]
\begin{center}
\includegraphics[width=\linewidth]{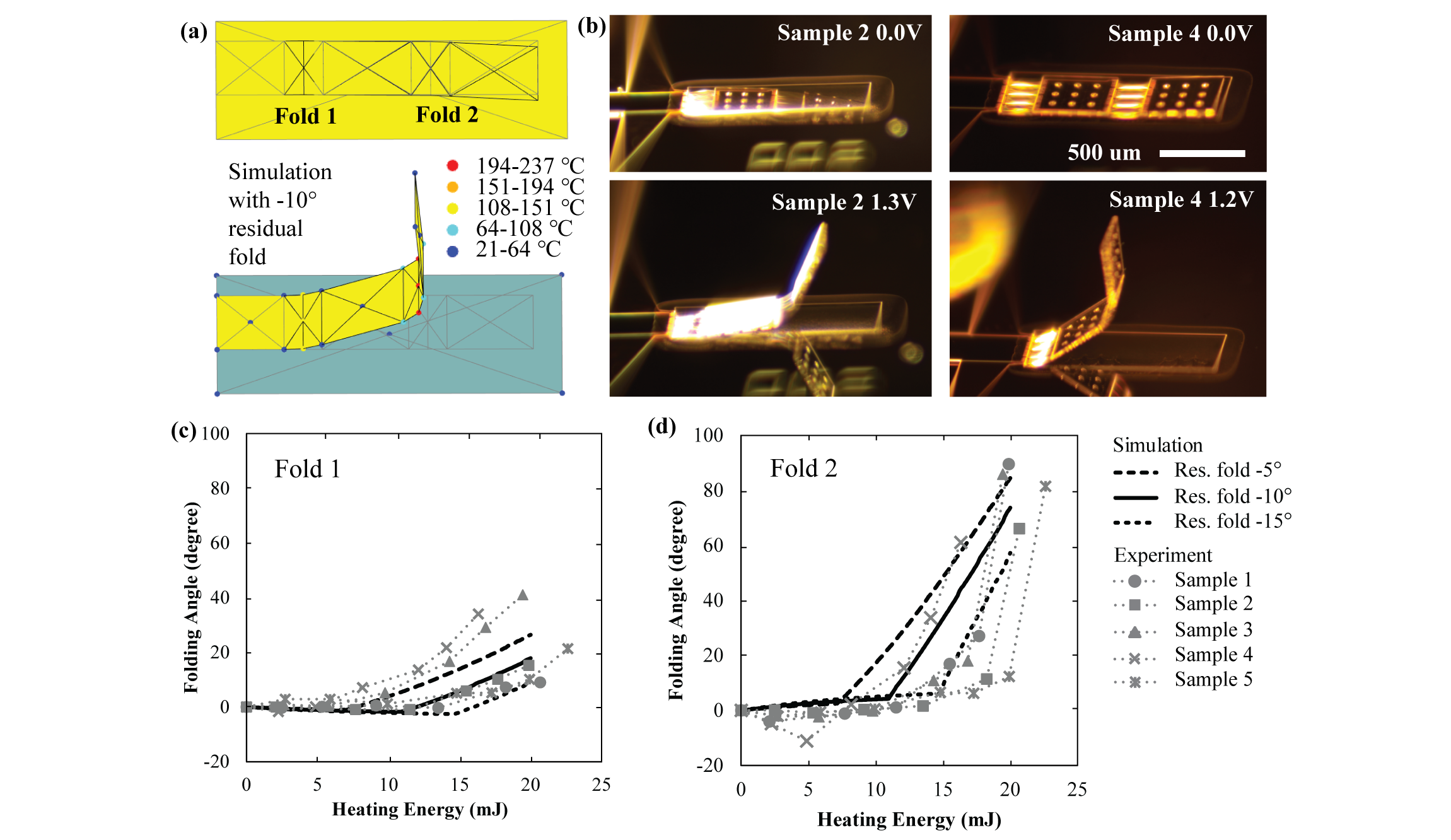}
\end{center}
\caption{Simulating the thermo-mechanically coupled folding in the double-crease origami  system. (a) Simulation of the folding shows that different temperatures are captured at the two creases; (b) Two experiments result in slightly different folding of the double-crease origami due to different residual stresses generated during the fabrication; (c-d) Folding curves showing the fold angle versus the input power for (c) Fold 1 and (d) Fold 2. See supplementary video Verification 2 for a comparison of the simulated animations and the recorded motion of the physical devices.}
\label{figure_double} 
\end{figure*}

First, we will study a single-crease micro-origami system which experiences a gravity induced folding threshold. The threshold effect occurs because gravity acting down on the origami causes it to come into contact with the substrate underneath. To fold the single crease origami, we need to apply power higher then a specific threshold to overcome the effects of gravity before observable folding is achieved (Fig.~\ref{figure_SingleCrease} (a)). This case study shows that the proposed model can accurately capture the multi-physics interaction between the heating induced self-folding, panel contact, and gravity.

Figure~\ref{figure_SingleCrease} (b) demonstrates the simulated folding behavior with our bar and hinge framework, and Fig.~\ref{figure_SingleCrease} (c) shows the comparison between the simulation and experimental measurements. The physical devices were fabricated and actuated with processes introduced in our previous work \cite{Zhu2020AFM}. The simulation curves are plotted with an assumed negative residual folding angle. This negative residual folding angle is a physical characteristic of the systems which occurs because of a residual stress generated during the fabrication of the polymer (SU-8) layer and the gold layer. This effect is difficult to quantify analytically, so we assume that the creases will develop about $-10^{\circ}$ of residual folding, which matches the observation of testing beams with no panels \cite{Zhu2020AFM}. The simulation results for $0^{\circ}$ and  $-20^{\circ}$ of residual folding are also plotted for comparison. The simulation is able to capture the threshold effects and the large folding of the origami accurately. The wall-clock time for running the entire simulation with 250 thermal increments is 13 seconds recorded on a modern laptop with i7-8750H processor. The model under-predicts the nonlinearity in the folding and a higher heating power is needed to reach the high folding angles. The more extreme nonlinearity observed in the real experiments is likely due to the nonlinear material responses and the complex thermal boundary generated by the etch holes surrounding the single-crease origami. Previous researches have shown that the mechanical and thermal properties of SU-8 is nonlinear \cite{Robin2014,Xu2016}, which can lead to the mismatch in Fig.~\ref{figure_SingleCrease} (c). Also, the thermal boundary with a complex 3-D shape is simplified as a 2-D plane in the proposed model so the heating dissipation is not captured exactly. 

Figure~\ref{figure_SingleCrease} (c) also contains one outlier that we encountered during the physical testing (Sample 5). In this case, the specimen experienced a local defects during the fabrication which caused the wrinkling in the electro-thermal actuator crease and delayed the folding from happening (Fig.~\ref{figure_SingleCrease} (d)). Such localized behaviors cannot be captured with the proposed model as our model has limited degrees of freedom and is dedicated to only simulating the global response of origami. Although this outlier illustrates the limitation of our model, we believe it does not compromise the usefulness of the proposed model. More robust fabrication methods will prevent such outliers from happening in the future.

\subsection{Double-crease origami with thermo-mechanical coupling}

Here, we study the thermo-mechanically coupled folding behavior within a double-crease origami pattern (Fig.~\ref{figure_double}). This system has two identical creases, however, the folding of the two creases is different because each crease has different thermal boundaries that results in different temperatures when heated. The second fold (fold 2 on Fig.~\ref{figure_double} (a)) is folded further away from the substrate which leads to a slower heat dissipation, a higher temperature, and a higher curvature. This coupled thermo-mechanical effect can be captured accurately with the proposed model. 

Figure~\ref{figure_double} (a) shows the simulated folded geometry of the origami  system, and Fig.~\ref{figure_double} (b) shows the experimental pictures of the same systems. The physical devices are fabricated using the processes reported in \cite{Zhu2020AFM}. A single simulation of the double-crease origami pattern takes about 18 seconds to complete (recorded on a modern laptop with i7-8750H processor). Figure~\ref{figure_double} (c) and (d) show a comparison between the simulation and the measured results for the folding angle of the two creases. Similar to the single-crease, the double-crease system also experiences the gravity induced threshold effect, which can be captured with the  bar and hinge formulation accurately. More significantly, the results show that the proposed model can capture the thermo-mechanical interaction during the large folding deformation. The simulation successfully predicts that the fold 2 will develop a higher temperature and a larger folding angle as it moves away from the substrate. 

The experimental results show that the fabricated physical samples have different residual folding as a result of the fabrication process (in Fig.~\ref{figure_double} (b), the two samples rest at different configurations when no heating is applied). The difference in residual folding is mainly caused by the variation in residual stresses of the SU-8 layer and the gold layer. This residual folding can be measured and estimated using the testing beam samples from our previous work\cite{Zhu2020AFM}. These variations in residual stress are common for micro-scale fabrication but they are tremendously difficult to control \cite{Kang2019}. We envision that the proposed simulation framework can be used to enumerate different combinations of design parameters efficiently to make the origami design robust against the influence of residual stress and other factors. Because the proposed framework is computationally efficient, it is possible to enumerate a large number of combinations of design parameters within a reasonable time schedule.

\begin{figure}[t]
\begin{center}
\includegraphics[width=\linewidth]{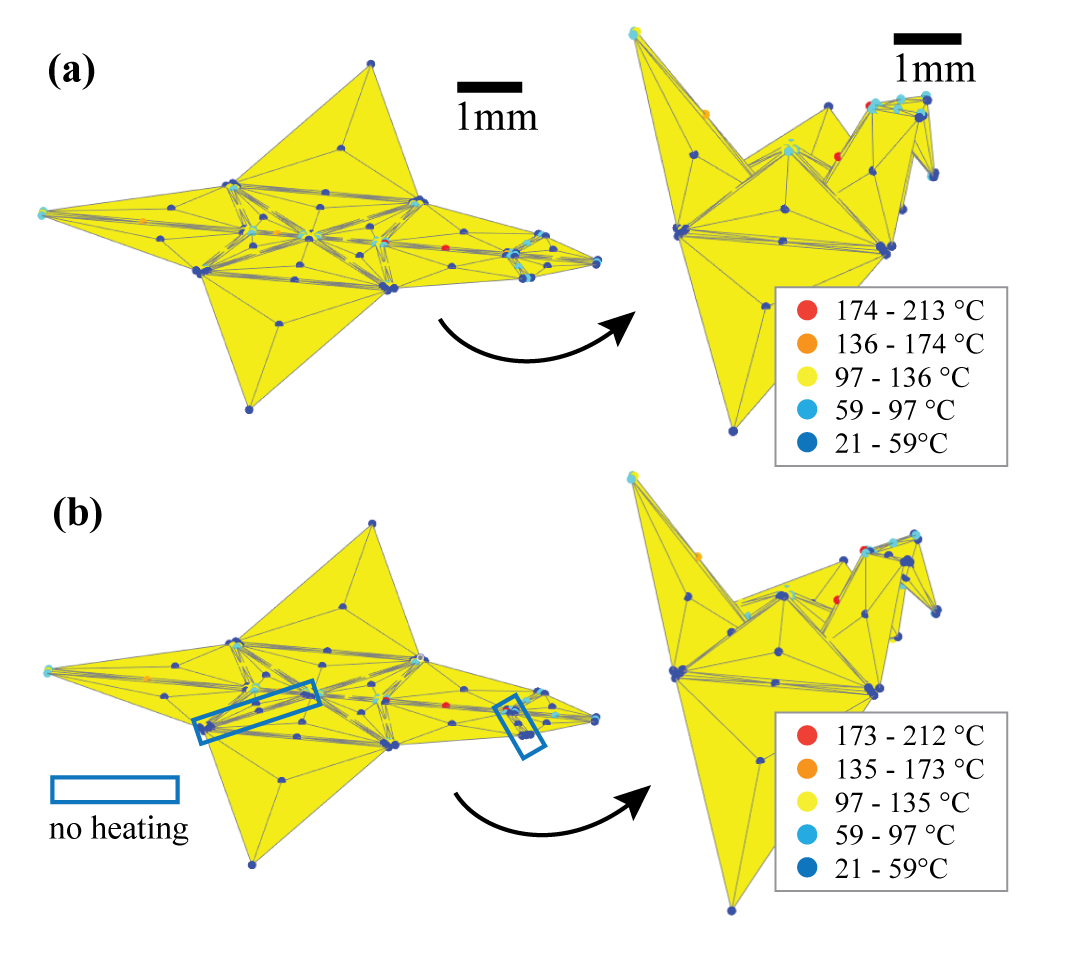}
\end{center}
\caption{\textcolor{black}{Simulation of the self-folding of an origami crane. (a) Folding the crane pattern with applied heating to all active creases. (b) Folding the crane pattern with two creases deactivated. Although not all of the creases are active, the crane can still fold successfully because the center pattern has a one-degree-of-freedom folding kinematic motion. This crane pattern has 17 creases and the simulation can be done in about 30 seconds recorded on a modern desktop with i9-10900K processor. See supplementary video Example 1 for the simulated animation.}}
\label{figure_ori} 
\end{figure}

\section{Application Examples}

In this section, we will demonstrate the effectiveness and efficiency of the proposed method with three application examples: simulation of a crane pattern, folding of a Miura origami lifter, and an optimization study of an origami micro-gripper. 

\subsection{Simulation of a crane pattern}

In the first example, we simulate the folding of an origami crane pattern as shown in Fig.~\ref{figure_ori}. This crane pattern has 17 creases and is more complex then the single-crease and double-crease systems demonstrated previously. We use this example to demonstrate that the provided implementation code package can be easily adapted to simulate more complex origami systems. We assume that the crane pattern is made with the same fabrication method as discussed previously \cite{Zhu2020AFM} and thus the panel are made of thick SU-8 and the electro-thermal creases are made with Au and thin SU-8 films. The folding is accomplished by applying heating power to all creases of the crane pattern simultaneously. For simplicity, we neglect the effects of gravity and the thermal boundary of the substrate in this example. 

\textcolor{black}{Figure~\ref{figure_ori} (a) demonstrates the folded geometry of this crane pattern, and the wall-clock time for running this simulation is about 30 seconds (recorded on a modern desktop with i9-10900K processor). When folding complex origami patterns, the folding speed of different creases tend to be different because of kinematic constraints by the pattern geometry \cite{Zhu2020JMR}. However, common fabrication methods usually cannot ensure that different creases will fold with different speeds \cite{Zhu2020AFM,Na2015}. Thus, there will be interference between the different folding creases and strain energy will develop in the creases as their fold angles deviate from the computed stress-free state \cite{Zhu2020JMR}. The proposed method can be used to study and simulate this interference effectively as demonstrated in Figure~\ref{figure_ori} (a). In general, by designing an origami pattern that has one-degree-of-freedom kinematic motion, one can also achieve the desired folding even if the actuation speed deviates from the kinematic folding speed.  Moreover, it is also possible that real fabrication cannot actuate all creases simultaneously, and there may be a need to leave certain creases as passive rotational springs \cite{Zhu2020AFM}. Figure~\ref{figure_ori} (b) demonstrates one example of non-uniform actuation where two creases of the crane pattern are deactivated. The simulation shows that although we are not applying heating to these two creases, the folding can still proceed successfully. The successful folding is achieved because the pattern of this crane has one-degree-of-freedom folding kinematics. The two computation examples show that the proposed simulation framework is well suited for studying the complex folding motion of active origami. Moreover, this example shows that the provided execution package  can be easily adapted to efficiently simulate the electro-thermal folding of relatively complicated origami systems with non-uniform actuator placement. The relevant simulation code ``\texttt{Example06\_Crane.m}'' can be found on the GitHub page.}

\begin{figure}[t]
\begin{center}
\includegraphics[width=\linewidth]{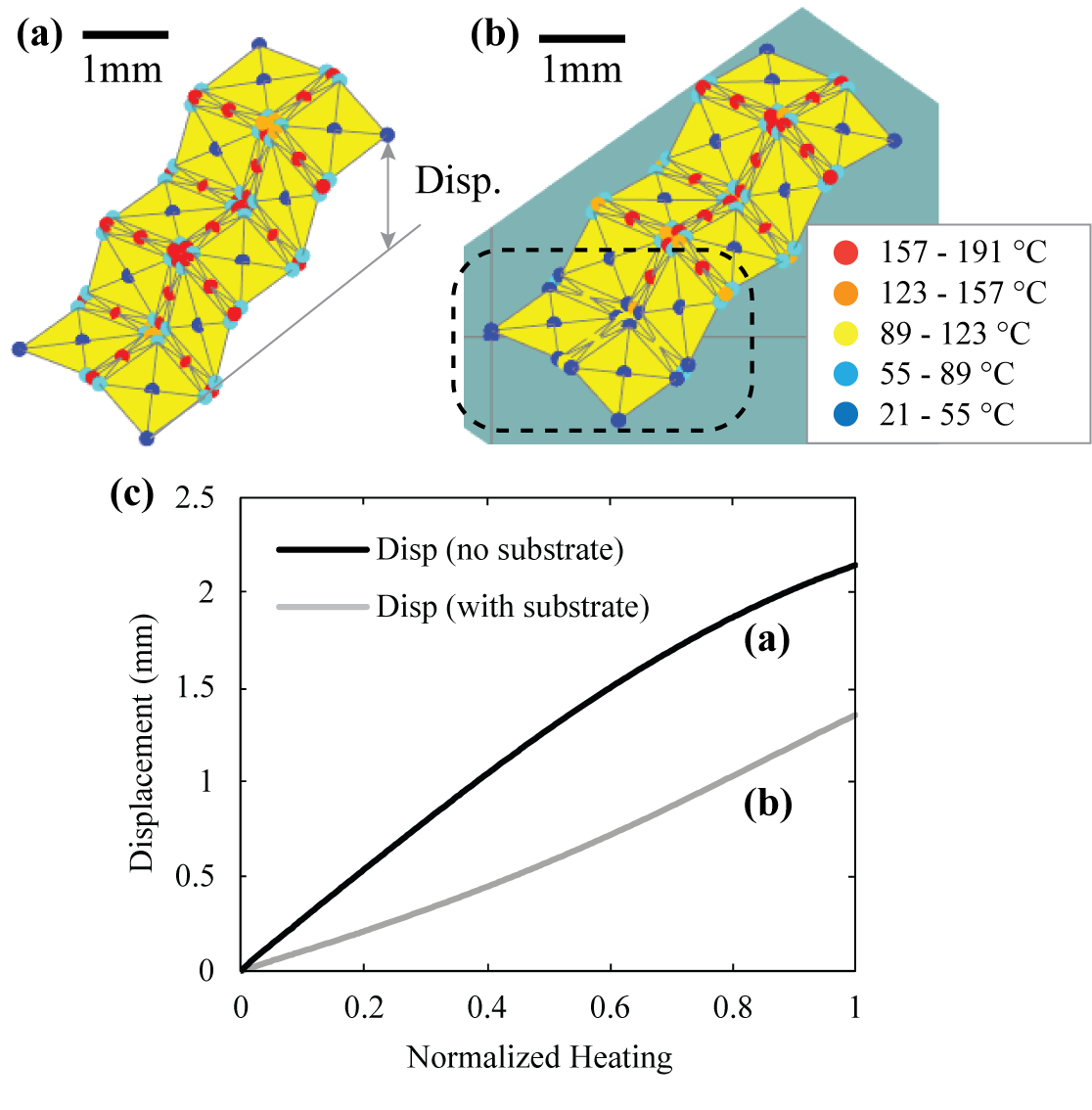}
\end{center}
\caption{Simulation of a Miura-ori pattern lifting device. (a) Deformed geometry of the lifting device when no substrate is presented; (b) Deformed geometry of the lifting device with a substrate underneath; (c) Comparison of the lifting distance with respect to the input heating for the two cases. See supplementary video Example 2 for the simulated animation.}
\label{figure_miura} 
\end{figure}

\begin{figure*}[t]
\begin{center}
\includegraphics[width=0.9\linewidth]{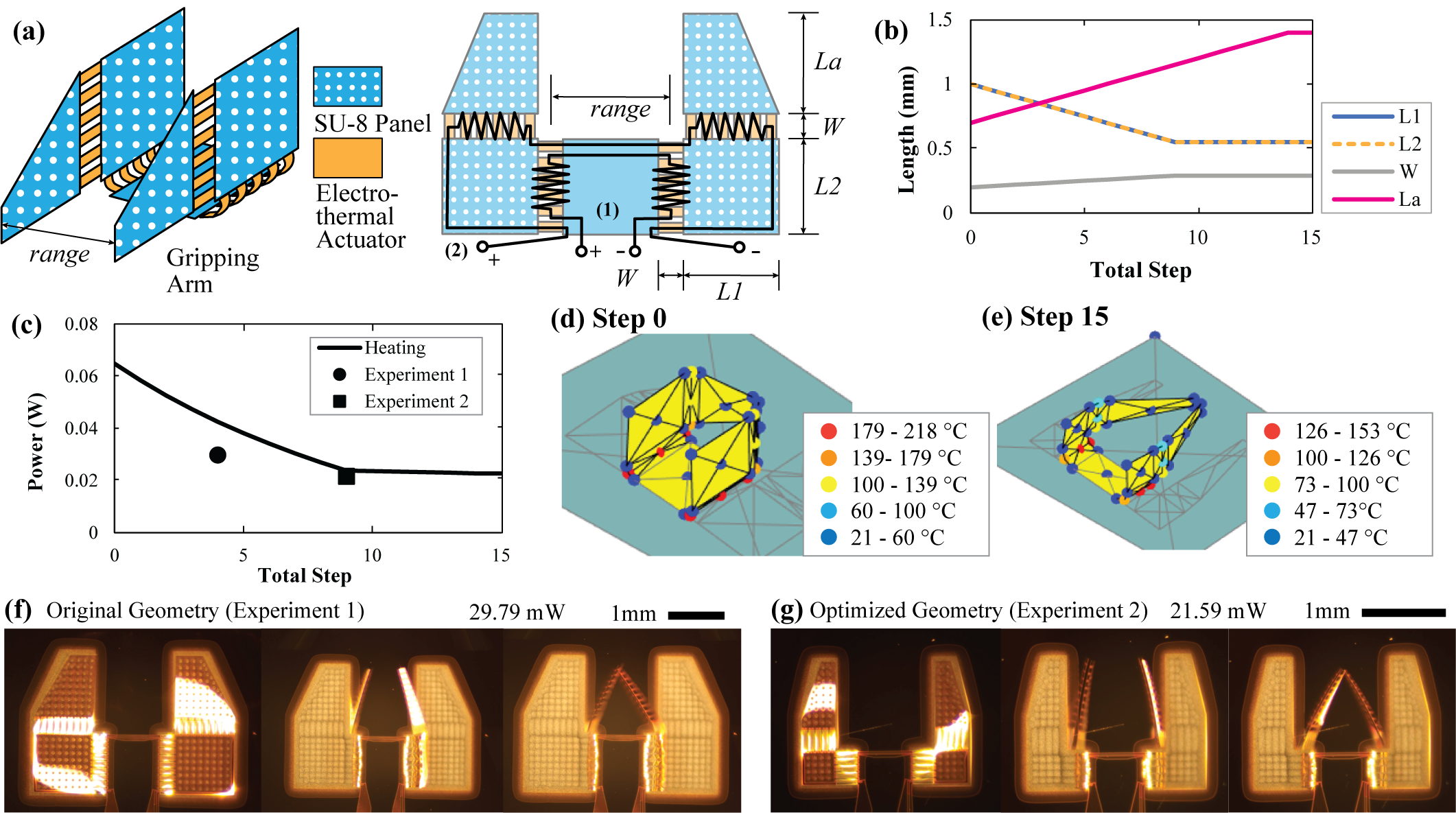}
\end{center}
\caption{Optimizing the performance of a micro-origami 3-D gripper. (a) The origami gripper configuration and dimensions before optimization; (b) The change in origami gripper dimensions during the optimization; (c) The change in power consumption during the optimization; (d) The simulated gripper geometry at step 0; (e) The simulated gripper geometry at step 15. (f) A physical gripper with a geometry similar to step 4 of the optimization; (g) A physical gripper with a geometry similar to the step 9 of the optimization. See supplementary video Example 3 for a comparison of the simulated geometry and the recorded actuation of the physical device.}
\label{figure_opt} 
\end{figure*}

\subsection{Folding of a Miura beam lifter}

In this second example, we explore the folding of a Miura beam system such that it can be used as a lifting device. We use this example to show that the proposed simulation framework can capture complex thermo-mechanical interactions even when simulating a relatively complex origami pattern. Figure~\ref{figure_miura} shows the loading set up of the Miura beam, where the left end of the beam is anchored. All creases of the pattern are heated simultaneously so that the right end of the beam will be lifted as the pattern folds. We first consider folding the device when no substrate is presented under it. Figure~\ref{figure_miura} (a) shows the folded the geometry when no supporting substrate is presented and the dark line in Fig.~\ref{figure_miura} (c) gives the lifting displacement with respect to the normalized heating power. 

This first loading set up is usually not realistic because a substrate is needed to support the device in a physical test. For small-scale devices, the large substrate supporting the devices will naturally introduce a temperature boundary condition. Figure~\ref{figure_miura} (b) shows the deformed geometry of the same Miura beam under the same electrical heat loading when the substrate is presented underneath. We can see that those creases on the left end of the beam (in the dashed box) have a lower temperature than those creases on the right end. This temperature reduction is observed because the creases on the left end are closer to the substrate and thus dissipate the heating power quicker. Because part of the creases develop lower temperature, the lifting displacement of the device is also lower compared to the situation when no substrate is presented (see Fig.~\ref{figure_miura} (c) gray line). A wall clock time of about 50 seconds is recorded to run the analysis of this Miura beam example on a modern laptop with i7-8750H processor, which shows the efficiency and the capability of capturing changing thermal boundary of the proposed simulation framework. The simulation code ``\texttt{Example03\_Miura.m}'' can be found on our GitHub.

\subsection{Optimization of origami gripper}

In this final example, we demonstrate how the simulation framework can be used as an efficient tool for optimizing the design of active origami devices for realistic application. \textcolor{black}{Over the past decades, there have been a number of research efforts to optimize origami designs \cite{Dudte2016}. For example, particle swarm optimization can be used for finding new origami tessellations \cite{Chen2021}, a mixed-integer-program can be used for determining the mountain-valley assignment of origami creases in a pattern \cite{Chen2020}, and genetic algorithms could be used for optimizing the geometry of origami patterns \cite{Gillman2019}. However, most of the existing optimization studies are applied onto kinematic simulations of origami systems and thus have limited capability to optimize performance based metrics of origami like the power consumption. With the proposed model, we now have the capability to simulate function related metrics of active origami efficiently and effectively, and thus can further achieve optimization of the performance of a functional origami.}

Here, the proposed framework is used to optimize the design of an active origami  gripper first conceived in \cite{Zhu2020AFM}. The original origami gripper demonstrated in our previous work \cite{Zhu2020AFM} was designed based on a trial and error approach and its performance was not optimized. This micro-gripper has four foldable panels controlled by two separate circuits (see Fig.~\ref{figure_opt} (a) for system design and dimensions). The first circuit is used to fold the base panels to assemble the gripper into the 3-D geometry, and the second circuit is used to close the arms to achieve gripping functions. 

In this example, we will search for a better geometrical design to minimize the input power needed to first assemble and then close the gripping arms. Practical constraints are considered and these include: (1) the size of the panels should not exceeds a prescribed range; (2) the gripping range is fixed as a constant; and (3) the temperature of the creases should not exceed the glass transition temperature of SU-8 so that the structure will not soften too much due to the excessive heating. The gripper optimization problem is formulated as follows:

\begin{flalign}
min \ & (Power(W,L_1,L_2,L_a)) \label{eq_opt1}
\\ 
s.t. \  &  L_{min} \leq L_1,L_2 \leq L_{max}  \label{eq_opt2}
\\ 
\ & W_{min} \leq  W \leq W_{max}  \label{eq_opt3}
\\ 
\ & L_{a,min} \leq L_a \leq L_{a,max} \label{eq_opt4}
\\ 
\ & max_i(T_{i}^{crease}) \leq 1.2\times T_{g,SU8} \label{eq_opt5} 
\end{flalign}

We use a coordinate descent with fixed step length to perform the optimization. The starting geometry is: $L_{1}^0=1000\mu m$, $L_{2}^0=1000\mu m$, $L_{a}^0=700\mu m$, and $W^0=200\mu m$, and the constraints are selected as: $L_{min}=500 \mu m$, $L_{max}=1500 \mu m$, $W_{min}=100 \mu m$, $W_{max}=300 \mu m$, $L_{a,min}=500 \mu m$, and $L_{a,max}=1500 \mu m$. Unlike the original design presented in \cite{Zhu2020AFM}, the panels of these micro-origami grippers are made with 5 $\mu m$ thick SPR photoresist rather than the 20 $\mu m$ thick SU-8 photoresist. This reduction in thickness makes the folding motion of the gripper easier to control by reducing effects of gravity. Figures~\ref{figure_opt} (b) to (e) summarize the results of the optimization. We can see that as the optimization proceeds, the base panels become smaller and the actuator creases become longer. These optimization results are obtained because the smaller base panels can reduce gravity loading and the longer actuator creases can achieve the same folding angle at a lower temperature with less input power. The optimized gripper not only demonstrates better energy performance, but also has lower functioning temperature, which makes it more robust and less likely to be overheated. Each simulation of the gripper can be done in 20 seconds and the entire optimization is accomplished in about 25 minutes without parallelization, demonstrating the high efficiency of the proposed simulation framework.

After running the simulation, the optimized gripper was fabricated and tested.  We recorded the folding motion and the power consumption as shown in Fig.~\ref{figure_opt} (f) and (g). The first gripper sample Fig.~\ref{figure_opt} (f) has a geometry that is close to the simulated gripper obtained in step 4 of the optimization, and the second gripper in Fig.~\ref{figure_opt} (g) has a geometry that is close to the simulated gripper obtained in step 9 of the optimization. The geometry of the fabricated samples does not match the simulated geometry exactly due to practical limitation of the fabrication. The first gripper design requires 29.8 mW to complete the full gripping function while the optimized sample requires 21.6 mW, demonstrating that the optimization successfully reduced the power consumption. However, the measured power consumption of the unoptimized gripper is lower than what is simulated. There are multiple factors that can lead to this overestimation. First of all, the residual curvature of the relatively thin SPR panel for the non-optimized gripper can interact with the folding process and contribute to the mismatch in the power calculation. Besides, a fixed heat dissipation angle $\beta$ is used for the entire system without considering the thermal interaction between adjacent creases. When there are multiple creases and panels that are close to each other (such as in this gripper), the heat dissipation angles $\beta$ should decrease to consider that the multiple creases/panels cannot dissipate as much heating power using a shared volume of air. Because the crease temperatures in the optimized design are substantially lower, there is less overall dissipation, and thus it matches simulated prediction better. Finally, variations in thickness, material properties, pattern sizes, and other factors during the fabrication could also contribute to the mismatch. Although the predicted power consumption does not match the the experiment perfectly, the revealed design principle is valid and  useful and the power consumption of the gripper is reduced successfully, which demonstrate capability of using the simulation framework for design optimization.

\section{Conclusion}

In this work, we proposed a novel simulation framework to capture the multi-physics behavior of electro-thermally actuated origami systems. The framework is based on the bar and hinge approach for representing origami, which allows for a simplified and rapid simulation of the active systems with thin-sheet geometry. This work introduces new formulations for capturing the heat transfer and the thermo-mechanical coupling within the origami. Our approach recursively solves (1) the heat transfer problem for the nodal temperature of the origami under an applied heating power, (2) the curvature of each crease from the elevated crease temperature, and (3) the new equilibrium position of the origami using a mechanical solver that determines the global folding motion of the system. 

In the proposed formulation, planar triangular thermal elements are used to capture heat transfer within the structure, while the structure to environment heat loss is simulated as a simplified 1-D heat conduction problem. This modeling technique is calibrated and verified against analytical solutions and FE simulations, and the result shows that the new model can accurately capture the two forms of heat transfer within active origami systems.

Next, this work presents two verification examples where we compare the predicted folding motion from the proposed simulation framework to physical experiments of electro-thermally actuated micro-origami. The two verification examples show that the simulation can capture the gravity induced folding threshold, the contact between the origami and a substrate, and the thermo-mechanically coupled folding motions. These two examples highlight the capability and validity of the proposed simulation framework as a rapid method to capture the behaviors of origami-inspired systems.

Finally, we presented three application examples where the model is used to study (1) the folding motion of a relatively complex origami crane pattern, (2) the influence of having a substrate under a Miura origami lifting device, and (3) the design optimization of an origami gripper. With these three examples, we further demonstrate the efficiency and effectiveness of the proposed framework for simulating the interdependent electro-thermal actuation, large deformation folding, contact between different panels, and other loading effects. 

This simulation framework provides a much needed method to simulate, design, and optimize electro-thermal or thermally actuated origami systems. Moreover, the proposed framework can be used at multiple length scales and is not limited to origami-type systems. By re-tuning the model parameters, we envision that the model can also be applied to simulate thermal and active multi-physical behaviors in other systems with planar features such as ribbon-based or kirigami-based structures. We believe that the proposed simulation framework will push the limits of origami systems, allowing engineers and researchers to explore new origami  systems, to create new control protocols for active origami structures, and to rapidly refine and optimize their designs. 

\section{Availability of the code package}

The code package with an implementation of the proposed simulation framework is available at: \url{https://github.com/zzhuyii/OrigamiSimulator}. The current implementation is in MATLAB. We are actively working on updating the package for including newer models and more advanced functionalities.

\section*{Acknowledgment}
The authors thank all staff members from the Lurie Nanofabrication Facility, especially K. Beach and P. Herrera-Fierro for their helpful advice and training on micro-fabrication. We also appreciate the helpful discussion and suggestions from Prof. K. Oldham and M. Birla at the University of Michigan. The authors acknowledge support from the Defense Advanced Research Project Agency (DARPA) Grant D18AP00071 and the National Science Foundation (NSF) Grant \#2054148. The first author would like to acknowledge College of Engineering Challenge Fellowship and from College of Engineering, University of Michigan. The paper reflects the views and position of the authors, and not necessarily those of the funding entities.

\bibliographystyle{elsarticle-num}
\bibliography{Reference}

\begin{thebibliography}{10}
\expandafter\ifx\csname url\endcsname\relax
  \def\url#1{\texttt{#1}}\fi
\expandafter\ifx\csname urlprefix\endcsname\relax\def\urlprefix{URL }\fi
\expandafter\ifx\csname href\endcsname\relax
  \def\href#1#2{#2} \def\path#1{#1}\fi

\bibitem{Felton2014}
S.~Felton, M.~Tolley, E.~Demaine, D.~Rus, R.~Wood, A method for building
  self-folding machines, Science 345~(6197) (2014) 644--646.
\newblock \href {https://doi.org/10.1126/science.1252610}
  {\path{doi:10.1126/science.1252610}}.

\bibitem{Rus2018}
D.~Rus, M.~T. Tolley, Design, fabrication and control of origami robots, Nature
  Reviews Materials 3 (2018) 101--112.
\newblock \href {https://doi.org/10.1038/s41578-018-0009-8}
  {\path{doi:10.1038/s41578-018-0009-8}}.

\bibitem{Sareh2020}
P.~Sareh, Y.~Chen, Intrinsic non-flat-foldability of two-tile ddc surfaces
  composed of glide-reflected irregular quadrilaterals, International Journal
  of Mechanical Sciences 185~(105881) (2020).
\newblock \href {https://doi.org/10.1016/j.ijmecsci.2020.105881}
  {\path{doi:10.1016/j.ijmecsci.2020.105881}}.

\bibitem{Overvelde2016}
J.~T. Overvelde, T.~A. DeJong, Y.~Shevchenko, S.~A. Becerra, G.~M. Whitesides,
  J.~C. Weaver, C.~Hoberman, K.~Bertoldi, A three-dimensional actuated
  origami-inspired transformable metamaterial with multiple degrees of freedom,
  nature communication 7~(10929) (2016).
\newblock \href {https://doi.org/10.1038/ncomms10929}
  {\path{doi:10.1038/ncomms10929}}.

\bibitem{Boatti2017}
E.~Boatti, N.~Vasios, K.~Bertoldi, Origami metamaterials for tunable thermal
  expansion, Advanced Materials 29~(1700360) (2017).
\newblock \href {https://doi.org/10.1002/adma.201700360}
  {\path{doi:10.1002/adma.201700360}}.

\bibitem{Silverberg2015}
J.~L. Silverberg, J.-H. Na, A.~A. Evans, B.~Liu, T.~C. Hull, C.~D. Santangelo,
  R.~J. Lang, R.~C. Hayward, I.~Cohen, Origami structures with a critical
  transition to bistability arising from hidden degrees of freedom, Nature
  Materials 14 (2015) 389--393.
\newblock \href {https://doi.org/10.1038/NMAT4232}
  {\path{doi:10.1038/NMAT4232}}.

\bibitem{Seymour2018}
K.~Seymour, D.~Burrow, A.~Avila, T.~Bateman, D.~C. Morgan, S.~P. Magleby, L.~L.
  Howell, Origami based deployable ballistic barrier, in: Origami7, Vol.~3,
  Proceedings of the 7th International Meeting on Origami in Science,
  Mathematics and Education, Oxford, UK, Sept 5-7, 2018, pp. 763--778.

\bibitem{Filipov2015}
E.~T. Filipov, T.~Tachi, G.~H. Paulino, Origami tubes assembled into stiff yet
  reconfigurable structures and metamaterials, PNAS 40~(112) (2015)
  12321--12326.
\newblock \href {https://doi.org/10.1073/pnas.1509465112}
  {\path{doi:10.1073/pnas.1509465112}}.

\bibitem{Lang2018book}
R.~Lang, Twists, Tilings, and Tessellations: Mathematical Methods for Geometric
  Origami, CRC Press, 2018.

\bibitem{Na2015}
J.-H. Na, A.~A. Evans, J.~Bae, M.~C. Chiappelli, C.~D. Santangelo, R.~J. Lang,
  T.~C. Hull, R.~C. Hayward, Programming reversibly self-folding origami with
  micropatterned photo-crosslinkable polymer trilayers, Advanced Materials 27
  (2015) 79--85.
\newblock \href {https://doi.org/10.1002/adma.201403510}
  {\path{doi:10.1002/adma.201403510}}.

\bibitem{NaurozePaulino2018}
S.~A. Nauroze, L.~S. Novelino, M.~M. Tentzeris, G.~H. Paulino, Continuous-range
  tunable multilayer frequency-selective surfaces using origami and inkjet
  printing, PNAS 115~(52) (2018) 13210--13215.
\newblock \href {https://doi.org/10.1073/pnas.1812486115}
  {\path{doi:10.1073/pnas.1812486115}}.

\bibitem{An2018}
B.~An, S.~Miyashita, A.~Ong, M.~T. Tolley, M.~L. Demaine, E.~D. Demaine, R.~J.
  Wood, D.~Rus, An end-to-end approach to self-folding origami structures, IEEE
  Transactions on Robotics 34~(6) (2018) 1409--1424.
\newblock \href {https://doi.org/10.1109/TRO.2018.2862882}
  {\path{doi:10.1109/TRO.2018.2862882}}.

\bibitem{Schenk2013}
M.~Schenk, S.~D. Guest, Geometry of miura-folded metamaterials, PNAS 110~(9)
  (2013) 3276--3281.
\newblock \href {https://doi.org/10.1073/pnas.1217998110}
  {\path{doi:10.1073/pnas.1217998110}}.

\bibitem{Fang2018}
H.~Fang, S.-C.~A. Chu, Y.~Xia, K.-W. Wang, Programmable self-locking origami
  mechanical metamaterials, Advanced Materials 30~(1706311) (2018).
\newblock \href {https://doi.org/10.1002/adma.201706311}
  {\path{doi:10.1002/adma.201706311}}.

\bibitem{Zhai2017}
Z.~Zhai, Y.~Wang, H.~Jiang, Origami-inspired, on-demand deployable and
  collapsible mechanical metamaterials with tunable stiffness, PNAS 115 (2017)
  2032--2037.
\newblock \href {https://doi.org/10.1073/pnas.1720171115}
  {\path{doi:10.1073/pnas.1720171115}}.

\bibitem{Li2019}
S.~Li, H.~Fang, S.~Sadeghi, P.~Bhovad, K.-W. Wang, Architected origami
  materials: How folding creates sophisticated mechanical properties, Advanced
  Materials 31~(1805282) (2019).
\newblock \href {https://doi.org/10.1002/adma.201805282}
  {\path{doi:10.1002/adma.201805282}}.

\bibitem{Lv2014}
C.~Lv, D.~Krishnaraju, G.~Konjevod, H.~Yu, H.~Jiang, Origami based mechanical
  metamaterials, Scientific Report 4~(5979) (2014).
\newblock \href {https://doi.org/10.1038/srep05979}
  {\path{doi:10.1038/srep05979}}.

\bibitem{Yang2017}
N.~Yang, J.~L. Silverberg, Decoupling local mechanics from large-scale
  structure in modular metamaterials, PNAS 114~(14) (2017) 3590--3595.
\newblock \href {https://doi.org/10.1073/pnas.1620714114}
  {\path{doi:10.1073/pnas.1620714114}}.

\bibitem{Lang2016}
R.~J. Lang, S.~Magleby, L.~Howell, Single degree-of-freedom rigidly foldable
  cut origami flashers, Journal of Mechanisms and Robotics 8~(031005) (2016).
\newblock \href {https://doi.org/10.1115/1.4032102}
  {\path{doi:10.1115/1.4032102}}.

\bibitem{Filipov2019}
E.~T. Filipov, G.~H. Paulino, T.~Tachi, Deployable sandwich surfaces with high
  out-of-plane stiffness, Journal of Structural Engineering 145~(04018244)
  (2019).
\newblock \href {https://doi.org/10.1061/(ASCE)ST.1943-541X.0002240}
  {\path{doi:10.1061/(ASCE)ST.1943-541X.0002240}}.

\bibitem{Kaddour2020}
A.-S. Kaddour, C.~A. Velez, M.~Hamza, N.~Brown, C.~Ynchausti, S.~P. Magleby,
  L.~Howell, S.~V. Georgakopoulos, A foldable and reconfigurable monolithic
  relfectarray for space application, IEEE Access 8 (2020) 219355--219366.
\newblock \href {https://doi.org/10.1109/ACCESS.2020.3042949}
  {\path{doi:10.1109/ACCESS.2020.3042949}}.

\bibitem{Zhu2020AFM}
Y.~Zhu, B.~Mayur, K.~R. Oldham, E.~T. Filipov, Elastically and plastically
  foldable electro-thermal micro-origami for controllable and rapid shape
  morphing, Advanced Functional Material 30~(2003741) (2020).
\newblock \href {https://doi.org/https://doi.org/10.1002/adfm.202003741}
  {\path{doi:https://doi.org/10.1002/adfm.202003741}}.

\bibitem{Breger2015}
J.~C. Breger, C.~Yoon, R.~Xiao, H.~R. Kwag, M.~O. Wang, J.~P. Fisher, T.~D.
  Nguyen, D.~H. Gracias, Self-folding thermo-magnetically responsive soft
  microgrippers, ACS Applied Materials and Interfaces 7~(5) (2015) 3398--3405.
\newblock \href {https://doi.org/10.1021/am508621s}
  {\path{doi:10.1021/am508621s}}.

\bibitem{Ma2014}
J.~Ma, Z.~You, Energy absorption of thin-walled square tubes with a prefolded
  origami pattern--part 1: Geometry and numerical simulation, Journal of
  Applied Mechanics 81~(011003) (2014).
\newblock \href {https://doi.org/10.1115/1.4024405}
  {\path{doi:10.1115/1.4024405}}.

\bibitem{Xiang2020}
X.~Xiang, G.~Lu, Z.~You, Energy absorption of origami inspired structures and
  materials, Thin-Walled Structures 157~(107130) (2020).
\newblock \href {https://doi.org/10.1016/j.tws.2020.107130}
  {\path{doi:10.1016/j.tws.2020.107130}}.

\bibitem{Kamrava2018}
S.~Kamrava, D.~Mousanezhad, S.~M. Felton, A.~Vaziri, Programmable origami
  strings, Advanced Materials Technologies 3~(1700276) (2018).
\newblock \href {https://doi.org/10.1002/admt.201700276}
  {\path{doi:10.1002/admt.201700276}}.

\bibitem{Liu2021}
Q.~Liu, W.~Wang, M.~F. Reynolds, M.~C. Cao, M.~Z. Miskin, T.~A. Arias, D.~A.
  Muller, P.~L. McEuen, I.~Cohen, Micrometer-sized electrically programmable
  shape-memory actuators for low-power microrobotics, Science Robotics
  6~(eabe6663) (2021).
\newblock \href {https://doi.org/10.1126/scirobotics.abe6663}
  {\path{doi:10.1126/scirobotics.abe6663}}.

\bibitem{Rogers2016}
J.~Rogers, Y.~Huang, O.~G. Schmidt, D.~H. Gracias, Origami mems and nems, MRS
  Bulletin 41 (2016) 123--129.
\newblock \href {https://doi.org/10.1557/mrs.2016.2}
  {\path{doi:10.1557/mrs.2016.2}}.

\bibitem{Zhu2020JMR}
Y.~Zhu, E.~T. Filipov, A bar and hinge model for simulating bistability in
  origami structures with compliant creases, Journal of Mechanisms and Robotics
  12~(021110) (2020).
\newblock \href {https://doi.org/10.1115/1.4045955}
  {\path{doi:10.1115/1.4045955}}.

\bibitem{Zhu2019PRSA}
Y.~Zhu, E.~Filipov, An efficient numerical approach for simulating contact in
  origami assemblages, Proceedings of the Royal Society A 475~(20190366)
  (2019).
\newblock \href {https://doi.org/10.1098/rspa.2019.0366}
  {\path{doi:10.1098/rspa.2019.0366}}.

\bibitem{Liu2017PRSA}
K.~Liu, G.~Paulino, Nonlinear mechanics of non-rigid origami: an efficient
  computational approach, Proceedings of Royal Society A 473~(20170348) (2017).
\newblock \href {https://doi.org/10.1098/rspa.2017.0348}
  {\path{doi:10.1098/rspa.2017.0348}}.

\bibitem{Yoon2014}
C.~Yoon, R.~Xiao, J.~Park, J.~Cha, T.~D. Nguyen, D.~H. Gracias, Functional
  stimuli responsive hydrogel devices by self-folding, Smart Materials and
  Structures 23~(094008) (2014).
\newblock \href {https://doi.org/10.1088/0964-1726/23/9/094008}
  {\path{doi:10.1088/0964-1726/23/9/094008}}.

\bibitem{Kang2019}
J.-H. Kang, H.~Kim, C.~D. Santangelo, R.~C. Hayward, Enabling robust
  self-folding origami by pre-biasing buckling direction, Advanced Material
  31~(0193006) (2019).
\newblock \href {https://doi.org/10.1002/adma.201903006}
  {\path{doi:10.1002/adma.201903006}}.

\bibitem{Leong2009}
T.~G. Leong, C.~L. Randall, B.~R. Benson, N.~Bassik, G.~M. Stern, D.~H.
  Gracias, Tetherless thermaobiochemically actuated microgrippers, PNAS 106
  (2009) 703--708.
\newblock \href {https://doi.org/10.1073/pnas.0807698106}
  {\path{doi:10.1073/pnas.0807698106}}.

\bibitem{Leong2008}
T.~G. Leong, Christina, L.~Randall, B.~R. Benson, A.~M. Zarafshar, D.~H.
  Gracias, Self-loading lithographically structured microcontainers: 3d
  patterned, mobile microwells, Lab on a Chip 8 (2008) 1621--1624.
\newblock \href {https://doi.org/10.1039/B809098J}
  {\path{doi:10.1039/B809098J}}.

\bibitem{Bassik2009}
N.~Bassik, G.~M. Stern, D.~H. Gracias, Microassembly based on hands free
  origami with bidirectional curvature, Applied Physics Letters 95~(091901)
  (2009).
\newblock \href {https://doi.org/10.1063/1.3212896}
  {\path{doi:10.1063/1.3212896}}.

\bibitem{Shaar2015}
N.~S. Shaar, G.~Barbastathis, C.~Livermore, Integrated folding, alignment, and
  latching for reconfigurable origami microelectromechanical systems, Journal
  of microelectromechanical systems 24 (2015) 1043--1051.
\newblock \href {https://doi.org/10.1109/JMEMS.2014.2379432}
  {\path{doi:10.1109/JMEMS.2014.2379432}}.

\bibitem{Iwase2005}
E.~Iwase, I.~Shimoyama, Multistep sequential batch assembly of
  three-dimensional ferromagnetic microstructures with elastic hinges, Journal
  of Microelectromechanical Systems 14 (2005) 1265--1271.
\newblock \href {https://doi.org/10.1109/JMEMS.2005.851814}
  {\path{doi:10.1109/JMEMS.2005.851814}}.

\bibitem{Yang2020}
X.~Yang, L.~Chang, N.~O. Perez-Arancibia, An 88-milligram insect-scale
  autonomous crawling robot driven by a catalytic artificial muscle, Science
  Robotics 5~(eaba0015) (2020).
\newblock \href {https://doi.org/10.1126/scirobotics.aba0015}
  {\path{doi:10.1126/scirobotics.aba0015}}.

\bibitem{Jager2000}
E.~W.~H. Jager, O.~Inganas, I.~Lundstrom, Microrobots for micrometer-size
  objects in aqueous media: Potential tools for single-cell manipulation,
  Science 288 (2000) 2335--2338.
\newblock \href {https://doi.org/10.1126/science.288.5475.2335}
  {\path{doi:10.1126/science.288.5475.2335}}.

\bibitem{Tachi2009Rigid}
T.~Tachi, Simulation of rigid origami, in: R.~J. Lang (Ed.), Origami 4, Decmber
  8-10 2015, Caltech, Pasadena, CA, CRC Press., 2009, pp. 175--187.

\bibitem{Tachi2010}
T.~Tachi, Geometric considerations for the design of rigid origami structures,
  in: Proceedings of the International Association for Shell and Spatial
  Structures (IASS) Symposium, November 8-12 2010, Shanghai, China.
  International association for shell and spatial structures., 2010.

\bibitem{Liu2016Merlin}
K.~Liu, G.~Paulino, Merlin: A matalab implementation to capture highly
  nonlinear behavior of non-rigid origami, in: Proceedings of IASS Annual
  Symposium, September 26-30, 2016, Tokyo, Japan. International association for
  shell and spatial structures, 2016, pp. 1--10.

\bibitem{Schenk2010}
M.~Schenk, S.~D. Guest, Origami folding: A structural engineering approach, in:
  Origami5, Proceedings of 5OSME, July 13-17, 2010, Singapore, CRC press, 2010,
  pp. 293--305.

\bibitem{Tachi2009Quad}
T.~Tachi, Generalization of rigid foldable quadrilateral mesh origami, Journal
  of the International Association for Shell and Spatial Structures 50 (2009)
  173--179.

\bibitem{Kawasaki1988}
T.~Kawasaki, M.~Yoshida, Crystallographic flat origamis, Memoirs of the Faculty
  of Science, Kyushu University 42~(2) (1988) 153--157.
\newblock \href {https://doi.org/10.2206/kyushumfs.42.153}
  {\path{doi:10.2206/kyushumfs.42.153}}.

\bibitem{Hawkes2010}
E.~Hawkes, B.~An, N.~M. Benbernou, H.~Tanaka, S.~Kim, E.~D. Demaine, D.~Rus,
  R.~J. Wood, Programmable matter by folding, Proceedings of the National
  Academy of Science 107~(28) (2010) 12441--12445.
\newblock \href {https://doi.org/10.1073/pnas.0914069107}
  {\path{doi:10.1073/pnas.0914069107}}.

\bibitem{Liu2018Merlin2}
K.~Liu, G.~Paulino, Highly efficient structural analysis of origami assemblages
  using the merlin2 software, in: Origami 7, the 7th International Meeting on
  Origami in Science, Mathematics and Education (7OSME), September 5-7, 2018,
  Oxford University, United Kingdom. Tarquin., 2018.

\bibitem{Filipov2017}
E.~Filipov, K.~Liu, M.~Schenk, G.~Paulino, Bar and hinge models for scalable
  analysis of origami, International Journal of Solids and Structures 124~(1)
  (2017) 26--45.
\newblock \href {https://doi.org/10.1016/j.ijsolstr.2017.05.028}
  {\path{doi:10.1016/j.ijsolstr.2017.05.028}}.

\bibitem{Zhu2019IDETC}
Y.~Zhu, E.~Filipov, Simulating compliant crease origami with a bar and hinge
  model, in: IDETC/CIE, no. DETC2019-97119, August 16-19, 2019, Anaheim, CA,
  USA. ASME, 2019.
\newblock \href {https://doi.org/10.1115/DETC2019-97119}
  {\path{doi:10.1115/DETC2019-97119}}.

\bibitem{Ma2018}
J.~Ma, J.~Song, Y.~Chen, An origami-inspired structure with graded stiffness,
  International Journal of Mechanical Science 136 (2018) 134--142.
\newblock \href {https://doi.org/10.1016/j.ijmecsci.2017.12.026}
  {\path{doi:10.1016/j.ijmecsci.2017.12.026}}.

\bibitem{Hussein2016}
H.~Hussein, A.~Tahhan, P.~L. Moal, G.~Bourbon, Y.~Haddab, P.~Lutz, Dynamic
  electro-thermal-mechanical modelling of a u-shaped electro-thermal actuator,
  J. Micromech. Microeng. 26~(025010) (2016).
\newblock \href {https://doi.org/10.1088/0960-1317/26/2/025010}
  {\path{doi:10.1088/0960-1317/26/2/025010}}.

\bibitem{Ozsun2009}
O.~Ozsun, B.~E. Alaca, A.~D. Yalcinkaya, M.~Yilmaz, M.~Zervas, Y.~Leblebici, On
  heat transffer at microscale with implications for for micro actuator
  desing., J. Micromech. Microeng. 19~(045020) (2009).
\newblock \href {https://doi.org/10.1088/0960-1317/19/4/045020}
  {\path{doi:10.1088/0960-1317/19/4/045020}}.

\bibitem{Iamoni2014}
S.~Iamoni, A.~Soma, Design of an electro-thermally actuated cell microgripper,
  Microsyst. Technol. 20 (2014) 869--877.
\newblock \href {https://doi.org/10.1007/s00542-013-2065-8}
  {\path{doi:10.1007/s00542-013-2065-8}}.

\bibitem{Timoshenko1925}
S.~Timoshenko, Analysis of bi-metal thermostats, J.O.S.A. \& R.S.I. 11 (1925).

\bibitem{Pezzulla2015}
M.~Pezzulla, S.~A. Shilig, P.~Nardinocchi, D.~P. Holmes, Morphing of geometric
  composites via residual swelling, Soft Matter 11 (2015) 5812--5820.
\newblock \href {https://doi.org/10.1039/c5sm00863h}
  {\path{doi:10.1039/c5sm00863h}}.

\bibitem{Huang1999}
Q.-A. Huang, N.~K.~S. Lee, Analysis and design of polysilicon thermal flexure
  actuator, J. Micromech. Microeng. 9 (1999) 64--70.
\newblock \href {https://doi.org/10.1088/0960-1317/9/1/308}
  {\path{doi:10.1088/0960-1317/9/1/308}}.

\bibitem{MicroChemSU8}
MicroChem, Su-8 permanent photoresists, Tech. rep., MicroChem (2019).

\bibitem{Robin2014}
C.~J. Robin, A.~Vishnoi, K.~N. Joonalagadda, Mechanical behavior and anisotropy
  of spin-coated su-8 thin films for mems, Journal of Microelectromechanical
  Systems 24~(1) (2014).
\newblock \href {https://doi.org/10.1109/JMEMS.2013.2264341}
  {\path{doi:10.1109/JMEMS.2013.2264341}}.

\bibitem{Xu2016}
T.~Xu, J.~H. Yoo, S.~Babu, S.~Roy, J.-B. Lee, H.~Lu, Characterization of the
  mechanical behavior of su-8 at microscale by viscoelastic analysis, J.
  Micromech. Microeng 26~(105001) (2016).
\newblock \href {https://doi.org/10.1088/0960-1317/26/10/105001}
  {\path{doi:10.1088/0960-1317/26/10/105001}}.

\bibitem{Dudte2016}
L.~H. Dudte, E.~Vouga, T.~Tachi, L.~Mahadevan, Programming curvature using
  origami tessellations, Nature Materials 15 (2016) 583--588.
\newblock \href {https://doi.org/10.1038/NMAT4540}
  {\path{doi:10.1038/NMAT4540}}.

\bibitem{Chen2021}
Y.~Chen, J.~Yan, J.~Feng, P.~Sareh, Particle swarm optimization based
  metaheuristic design generation of non-trivial flat-foldable origami
  tessellations with degree-4 vertices, Journal of Mechanical Design
  143~(011703-1) (2021).
\newblock \href {https://doi.org/10.1115/1.4047437}
  {\path{doi:10.1115/1.4047437}}.

\bibitem{Chen2020}
Y.~Chen, L.~Fan, Y.~Bai, J.~Feng, P.~Sareh, Assigning mountain-valley fold
  lines of flat-foldable origami patterns based on graph theory and
  mixed-integer linear programming, Computers and Structures 239~(106328)
  (2020).
\newblock \href {https://doi.org/10.1016/j.compstruc.2020.106328}
  {\path{doi:10.1016/j.compstruc.2020.106328}}.

\bibitem{Gillman2019}
A.~S. Gillman, K.~Fuchi, P.~R. Buskohl, Discovering sequenced origami folding
  through nonlinear mechanics and topology optimization, Journal of Mechanical
  Design 141~(041401) (2019).
\newblock \href {https://doi.org/10.1115/1.4041782}
  {\path{doi:10.1115/1.4041782}}.

\end{thebibliography}

\end{document}